\documentclass[sigconf]{aamas} 
\usepackage{balance} % for balancing columns on the final page
\usepackage{graphicx}
\usepackage{subcaption}
\usepackage{xcolor}
\usepackage{multirow}
\usepackage{gensymb}
\usepackage{hyperref}
\usepackage{soul}

\setcopyright{ifaamas}
\acmConference[AAMAS '22]{Proc.\@ of the 21st International Conference
on Autonomous Agents and Multiagent Systems (AAMAS 2022)}{May 9--13, 2022}
{Online}{P.~Faliszewski, V.~Mascardi, C.~Pelachaud,
M.E.~Taylor (eds.)}
\copyrightyear{2022}
\acmYear{2022}
\acmDOI{}
\acmPrice{}
\acmISBN{}

\acmSubmissionID{754}

\title{ 
Intention-Aware Navigation in Crowds with Extended-Space POMDP Planning
}
\author{Himanshu Gupta}
\affiliation{
  \institution{University of Colorado Boulder}
  \city{Boulder}
  \state{Colorado}
  \country{USA}}
\email{himanshu.gupta@colorado.edu}

\author{Bradley Hayes}
\affiliation{
  \institution{University of Colorado Boulder}
  \city{Boulder}
  \state{Colorado}
  \country{USA}}
\email{bradley.hayes@colorado.edu}

\author{Zachary Sunberg}
\affiliation{
  \institution{University of Colorado Boulder}
  \city{Boulder}
  \state{Colorado}
  \country{USA}}
\email{zachary.sunberg@colorado.edu}

\acmSubmissionID{754}

\begin{abstract}
This paper presents a hybrid online Partially Observable Markov Decision Process (POMDP) planning system that addresses the problem of autonomous navigation in the presence of multi-modal uncertainty introduced by other agents in the environment.  
As a particular example, we consider the problem of autonomous navigation in dense crowds of pedestrians and among obstacles. Popular approaches to this problem first generate a path using a complete planner (e.g., Hybrid A*) with ad-hoc assumptions about uncertainty, then use online tree-based POMDP solvers to reason about uncertainty with control over a limited aspect of the problem (i.e. speed along the path). We present a more capable and responsive real-time approach enabling the POMDP planner to control more degrees of freedom (e.g., both speed AND heading) to achieve more flexible and efficient solutions. 
This modification greatly extends the region of the state space that the POMDP planner must reason over, significantly increasing the importance of finding effective roll-out policies within the limited computational budget that real time control affords. Our key insight is to use multi-query motion planning techniques (e.g., Probabilistic Roadmaps or Fast Marching Method) as priors for rapidly generating efficient roll-out policies for every state that the POMDP planning tree might reach during its limited horizon search. 
Our proposed approach generates trajectories that are safe and significantly more efficient than the previous approach, even in densely crowded dynamic environments with long planning horizons.

\end{abstract}


\newcommand{\BibTeX}{\rm B\kern-.05em{\sc i\kern-.025em b}\kern-.08em\TeX}

\begin{document}

%%% The following commands remove the headers in your paper. For final 
%%% papers, these will be inserted during the pagination process.

\pagestyle{fancy}
\fancyhead{}

%%% The next command prints the information defined in the preamble.

\maketitle 

%%%%%%%%%%%%%%%%%%%%%%%%%%%%%%%%%%%%%%%%%%%%%%%%%%%%%%%%%%%%%%%%%%%%%%%%

%%%%%%%%%%%%%%%%%%%%%%%%%%%%%%%%%%%%%%%%%%%%%%%%%%%%%%%%%%%%%%%%%%%%%%%%%%%%%%%%
\section{INTRODUCTION}

It is increasingly common to find autonomous vehicles operating successfully in relatively predictable and structured scenarios such as freeway driving.
However, despite ample investments, more complex navigation tasks with less structure imposed on the dynamic elements remain open challenges~\cite{kerry2017gauging,metz2021costly}.
Interaction with other agents in the environment is a particularly prolific source of difficult problems.
Navigating through a crowd of pedestrians is one important example of this.
In science fiction movies like Star Wars, droids move deftly between the people walking around them, and intuitively pedestrians should not greatly impede a properly-controlled robot's motion.
In order to choose a good trajectory, however, the robot must reason about the intentions of the humans around it, a task fraught with uncertainty.

\begin{figure}
    \centering
        \includegraphics[width=.85\columnwidth]{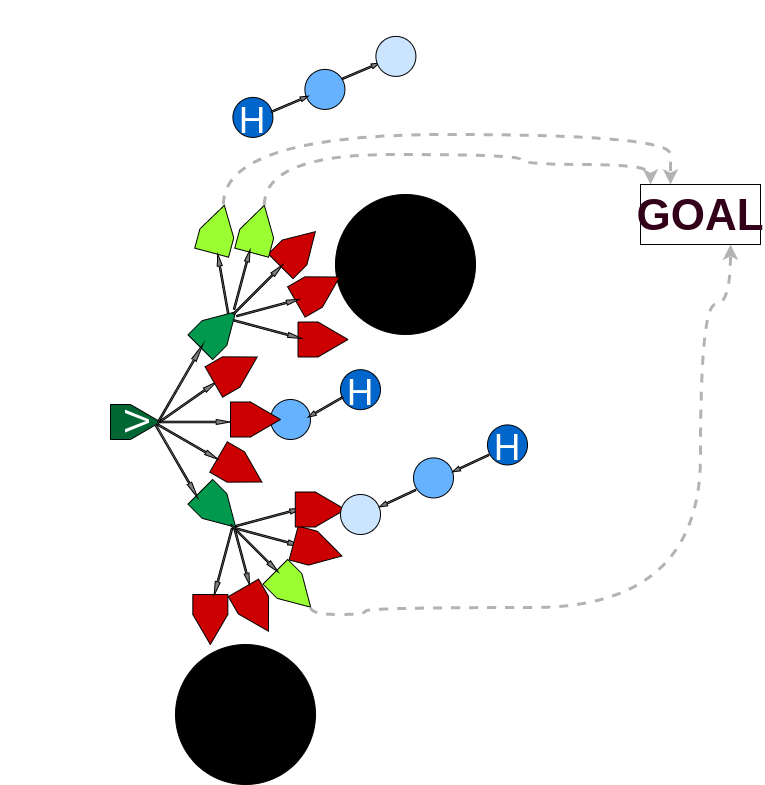}
        \caption{Two-dimensional POMDP motion planning with pedestrians. Green and red objects represent nodes in the planning tree, with green indicating high value. Blue circles denote the position of humans at different times. Black circles denote static obstacles. Dashed lines represent roll-out trajectories, a critical part of the proposed approach.}
        \label{explored_trajectories}
        \vspace{-6mm}
\end{figure}

The Partially Observable Markov Decision Process (POMDP) is a mathematical framework for optimal decision making in the presence of various types of uncertainty.
Previous approaches to tasks like pedestrian navigation have used the POMDP framework for navigation-centric tasks (e.g. \cite{bai2015intention,agha2014firm,bry2011rapidly,sunberg2017value,kim2018social,luo2018porca}).  
However, in these approaches POMDP planning is often relegated to a very limited role (e.g. speed control), or to a very limited class of uncertainty (e.g. Gaussian or unimodal distributions~\cite{agha2014firm,bry2011rapidly}).
For instance, Bai et al. (2015) \cite{bai2015intention} use the hybrid $A^*$ algorithm \cite{dolgov2008practical} to plan a drivable path from the vehicle's current position to its goal location, using a POMDP formulation only for speed control over that path.
The observed reticence within the field to use POMDPs for large planning problems is understandable; in general obtaining exact solutions to POMDPs is an intractable problem~\cite{papadimitriou1987complexity}.
However, sparse tree-based online planners are surprisingly insensitive to the size of the state and observation space~\cite{kearns2002sparse,silver2010pomcp,ye2017despot,sunberg2018pomcpow}, suggesting a way forward for increasingly expressive POMDP formulations to improve the state of the art in critical control problems.

In contrast to the partitioned approach, we propose a more effective and general approach to planning for real-time tasks involving navigation with partial observability and arbitrary distributions. We assert that the POMDP planner should have control over all degrees of freedom that are relevant to the uncertainty it is facing to maximize its ability to generate satisfactory plans. For example, when navigating among pedestrians, the POMDP planner should have control over the speed \emph{and heading}, rather than solely speed along a predetermined path \cite{bouton2017belief,hubmann2018automated,hubmann2017decision,bai2015intention}.
One may use online POMDP algorithms (e.g., DESPOT \cite{ye2017despot}) that perform a tree search guided by value estimates obtained by executing a roll-out policy. 

% In particular, we believe that the problem of planning among dynamic obstacles should be formulated as a 2-$D$ action space POMDP where the POMDP planner determines both, the best heading angle and speed at any moment of time. This POMDP can be solved online using existing algorithms. Online POMDP solvers like DESPOT \cite{ye2017despot} perform tree search that is guided by value estimates obtained by executing a roll-out policy. In the absence of an effective roll-out policy, the planner might never find the sparse positive rewards that are typical in navigation tasks.
% For a 2-$D$ action space POMDP planner, it is infeasible to use the same roll-out policy since generating a hybrid $A^*$ path from every possible future position of the vehicle during tree search to its goal location is computationally infeasible within the limited planning time.

Since an expansion of the action space can open up a much larger region of the state space to exploration, a critical challenge is determining a good roll-out policy for the vastly increased set of states reachable in the tree search\footnote{Since previous comparable approaches plan only along a 1-$D$ path generated via $A^*$, roll-out policies are needed only for that single path, which are straightforward to specify by hand~\cite{bai2015intention}}. In the absence of an effective roll-out policy, a limited-horizon planner might never find the sparse positive terminal rewards that are typical in navigation tasks. The proposed method addresses this by incorporating multi-query motion planning techniques to produce a more informed roll-out policy, allowing for a corresponding increase in POMDP complexity and thus solution quality. To demonstrate the effectiveness of this approach we evaluate it with two motion planning methods, Probabilistic Roadmaps ($PRM$) and Fast Marching Methods ($FMM$), to generate effective roll-out policies.

Our evaluation shows that online navigation solutions to a POMDP with an extended action space and roll-out policies informed by multi-query planning methods are considerably more efficient in densely crowded environments than the two-step approach proposed in \cite{bai2015intention} without compromising the safety of the pedestrians. The proposed approach explores multiple possible paths over a wider search space while reasoning over uncertainty in pedestrian intention as can be seen from Fig. \ref{explored_trajectories}, instead of handling uncertainty over just one path \cite{bai2015intention,bouton2017belief,hubmann2018automated,hubmann2017decision}. 
We have also shown that the choice of multi-query planning technique does not affect the performance significantly, as long as it can generate an effective roll-out policy. 
Using pedestrian navigation as a motivating example for our proposed method throughout the remainder of this work, we refer to the popular unidimensional speed-based POMDP control formulation as Limited Space planner or $LS$ planner, and our higher-dimensional (speed and heading) action space POMDP formulation as Extended Space planner or $ES$ planner.

\section{RELATED WORK}

In recent years, a number of research efforts have focused on solving the problem of autonomous navigation in dynamic environments, especially among pedestrians. 

For safe and efficient navigation, a controller should incorporate pedestrian intentions and the corresponding behaviors into decision making. This raises the need for accurate models of pedestrian intention and behavior. A considerable amount of work has focused on using recorded trajectories to learn pedestrian dynamics \cite{chen2016predictive,alahi2016social,mohamed2020social}. However, these methods generally have large data requirements, and the learned model may not generalize well to new conditions. Since these learned patterns generally do not change after their generation, Vasquez et al. \cite{vasquez2009growing} presented an approach where motion patterns can be learned incrementally, and in parallel with prediction using Hidden Markov Models. Luo et al.\cite{luo2018porca} designed a pedestrian motion model that accounts for both intentions and interactions to capture pedestrian motions accurately.
% and used POMDP planning to incorporate their uncertainties in a principled manner.  

Our work can be situated within a body of literature that focuses on determining the best plan of action for an autonomous vehicle, given a pedestrian behavior model. Less complex approaches use reactive control schemes (\cite{schiller1998collision, fox1997dynamic}) that neither utilize a pedestrian model nor account for the delayed effects of the agent's current action. As a result, these approaches often lead to sub-optimal decisions. Another common approach is to use deterministic pedestrian behavior models to generate paths that avoid dynamic and static obstacles. The path can subsequently be executed using a feedback controller (\cite{van2011lqg, kuwata2009real}). However, both approaches ignore the uncertainty in pedestrian intention estimation. Recent work has addressed this issue by formulating the problem as a POMDP, and then solving it by either using techniques from deep reinforcement learning (RL) \cite{chen2017decentralized,schulman2017proximal} or online POMDP solvers \cite{ye2017despot,cai2021hyp,leemagic,sun2020stochastic}. 
% Online POMDP planning generates a plan by reasoning over state uncertainty, estimating the value of various action sequences whose value is conditioned on the observations they may produce. It is effective in navigation tasks because it can reason over not just the uncertainty in pedestrian intention estimate, but also uncertainties in vehicle control and sensing in its decision making process. 

The works most closely related to this paper use online POMDP planning for this task of navigating under uncertainty\cite{bai2015intention,sunberg2017value,luo2018porca,song2016intention,bouton2017belief,hubmann2018automated,hubmann2017decision}. Bai et. al \cite{bai2015intention} tackled the complex task of navigation among pedestrians using a two step process. They used hybrid $A^*$ to obtain a sequence of steering angles that can guide the vehicle to its goal, and then used a POMDP planner which reasoned over the uncertainty in nearby pedestrians' intention to control the speed over that path. This two step process can lead to undesirable stalling of the vehicle. 
% Also, as hybrid $A^*$ needs a fixed initial location for path planning, this approach can not reason over uncertainty in vehicle's position. 
Luo et. al \cite{luo2018porca} compared their planner's performance against a dynamic hybrid $A^*$ approach that planned over both heading angle and speed, and the dynamic hybrid $A^*$ approach  outperformed other planners in all of their evaluation metrics. However, it led to collisions with pedestrians because unlike POMDP planning, dynamic hybrid $A^*$ path planning does not have the capability to handle pedestrian intention uncertainty. 
% Increasing the action space for hybrid $A^*$ planning comes at additional computation cost, which becomes more substantial for long horizon tasks. 
MAGIC \cite{leemagic} showed the effectiveness of using macro actions, a combination of both steering and speed in POMDP planning for autonomous driving in crowded environments. This suggests that the POMDP planner should control both degrees of freedom for safe and efficient planning. However, MAGIC introduces an additional step of learning macro actions which, we argue, is not necessary if suitable roll-out policies are used. 

Liang et al. \cite{liang2020realtime}, Sathyamoorthy et al. \cite{sathyamoorthy2020densecavoid} and Fan et al. \cite{long2018towards} used PPO \cite{schulman2017proximal} to train an RL policy that directly maps sensor data to vehicle velocity for collision avoidance with dynamic obstacles. However, these RL agents are hard to train for long range navigation tasks in complex environments where reward is sparse \cite{prm_rl}. SA-CADRL \cite{chen2017socially} uses a global planner \cite{chen2016motion} to generate way-points/sub-goals in close proximity, and used an RL planner to obtain socially acceptable collision free path between those way-points. To solve long range navigation tasks with just static obstacles, PRM-RL \cite{prm_rl} uses motion planning techniques, primarily sampling based methods for generating a roadmap using the RL agent to determine connectivity, rather than the traditional collision free straight line interpolation in C-space. RL-RRT \cite{rl_rrt} applies similar idea but also imposed kinodynamic constraints on the local RL planner. They showed the effectiveness of offline methods in guiding the optimal decision search.
% and incites their use in the presence of dynamic obstacles as well. 

Our work shows that online POMDP planning over increased degrees of freedom is achievable and more effective than controlling only a subset, without the need to learn and incorporate macro actions~\cite{leemagic}. The advantages of expanded space POMDP planning comes at the cost of higher computational complexity which can be offset by the use of offline methods\footnote{In practice, the "offline" portion of the computation can be carried out online, but at a slower rate than the POMDP planning.}. The purpose of this work is to demonstrate that POMDP planning is an effective tool that is capable of rapidly solving large-horizon planning problems, provided it can be guided by effective roll-out policies. To the best of the authors' knowledge, this is the first work that combines POMDP planning over multiple degrees of freedom with multi-query motion planning approaches for real time navigation in continuous dynamic environments with multi-modal process uncertainty. 

% \section{The Preamble}

% You will be assigned a submission number when you register the abstract 
% of your paper on \textit{EasyChair}. Include this number in your 
% document using the `\verb|\acmSubmissionID|' command.

% Then use the familiar commands to specify the title and authors of your
% paper in the preamble of the document. The title should be appropriately 
% capitalised (meaning that every `important' word in the title should 
% start with a capital letter). For the final version of your paper, make 
% sure to specify the affiliation and email address of each author using 
% the appropriate commands. Specify an affiliation and email address 
% separately for each author, even if two authors share the same 
% affiliation. You can specify more than one affiliation for an author by 
% using a separate `\verb|\affiliation|' command for each affiliation.

% Provide a short abstract using the `\texttt{abstract}' environment.
 
% Finally, specify a small number of keywords characterising your work, 
% using the `\verb|\keywords|' command. 

%%%%%%%%%%%%%%%%%%%%%%%%%%%%%%%%%%%%%%%%%%%%%%%%%%%%%%%%%%%%%%%%%%%%%%%%
\section{TECHNICAL APPROACH} \label{sec:approach}

This section describes the different technical components of our approach including our POMDP model, the DESPOT algorithm, and the multi-query planning techniques that underpin its performance.

\subsection{POMDP Preliminaries} 
% Define abbreviations and acronyms the first time they are used in the text, even after they have been defined in the abstract. Abbreviations such as IEEE, SI, MKS, CGS, sc, dc, and rms do not have to be defined. Do not use abbreviations in the title or heads unless they are unavoidable.

% A POMDP is a generalization of a Markov decision process (MDP). Both MDP and POMDP are mathematical frameworks for representing a wide range of sequential decision making problems. In the POMDP framework, the system dynamics are determined by an MDP, but the agent cannot directly observe the underlying state. Instead, it must maintain a probability distribution over the set of possible states, based on a set of observations and observation probabilities, and the underlying MDP. 

The Markov Decision Process (MDP) is a mathematical framework for representing a broad class of sequential decision making problems. A POMDP is a generalization of an MDP in which the agent cannot directly observe the underlying state. Instead, it must maintain a probability distribution over the set of possible states, based on a set of observations and observation probabilities, and the underlying MDP.

A POMDP is defined by a tuple $(S, A, Z, T, O, R, \gamma)$, where $S$ is the state space, $A$ is the action space, $Z$ is the observation space, $T$ is the transition model, $O$ is the observation model, $R$ is the reward model, and $\gamma$ is the discount factor.
When the system is in state $s \in S$ and takes an action $a \in A$, it reaches state $s' \in S$ with probability $T(s,a,s')$ and gets an observation $z \in Z$ with probability $O(s',a,z)$. The reward model $R$ is specified by a function $R(s,a,s')$ which specifies the immediate reward of transitioning from state $s$ via action $a$ to state $s'$.

% One method for handling the lack of direct state observability is to maintain a belief over all the possible states. Let $b_{t-1}$ be the belief at time $t-1$. If the system takes an action $a_t$ and gets an observation $z_t$ at the next time step $t$, then using Bayes' rule, we get the new belief $b_t$ as:

%\begin{equation} \label{eqn_1}
%b_t(s') = \eta O(s', a_t, z_t) \sum_{s\in S} T(s, a_t, s') b_{t-1}(s)
%\end{equation}

%where $\eta$ is a normalization constant.

A policy for a POMDP is a function $\pi$ that specifies the action $a = \pi(b)$ at any given belief over the state space $b$. Online POMDP solvers generate a policy that maximizes the expected total reward from the current belief $b$: 

\begin{equation} \label{eqn_2}
V_{\pi}(b) = \mathrm{E}\big(\sum_{t=0}^{\infty} \gamma^t R(s_t, \pi(b_t)\big)|b_0 = b)
\end{equation}

% where $\gamma \in (0,1)$ is a discount factor, which specifies the preference of immediate goals over future goals.
% BH: Gamma is explained in the POMDP definition

\subsection{Problem formulation as a POMDP} 
% Define abbreviations and acronyms the first time they are used in the text, even after they have been defined in the abstract. Abbreviations such as IEEE, SI, MKS, CGS, sc, dc, and rms do not have to be defined. Do not use abbreviations in the title or heads unless they are unavoidable.

Our approach utilizes a POMDP to model both the agent and the dynamic obstacles around it, generating control solutions that account for uncertainty in the environment.

\subsubsection{State Modeling}

The state vector in our dynamic environment navigation task POMDP consists of the vehicle state and a vector of dynamic obstacle states. The vehicle state consists of position $(x_c,y_c)$, orientation $\theta_c$, current speed $v_c$ and its goal location $g_c$.
The state vector contains $n_{ped}$ pedestrian states whose future motion intentions are not directly observable, contributing uncertainty in the problem formulation. The state of the $i^{th}$ pedestrian consists of its position $(x_i, y_i)$, speed ${v_i}$, and its intended goal location $g_i$. The intention of a pedestrian is modeled as a goal location, which is hidden from the vehicle and must be inferred from its observed behavior.

\subsubsection{Action Modeling}

The action space in the navigation POMDP consists of a two dimensional vector where the agent chooses both steering (the change in the orientation angle, $\delta_{\theta}$) and velocity (the change in vehicle's speed, $\delta_{s}$) controls, with the range of possible values for each being dependent on the vehicle state.
Further details on this are available in Section \ref{sec:diff_planners}. There is also a SUDDEN BRAKE $(SB)$ action that immediately stops the vehicle to avoid collision with pedestrians in unexpected scenarios. 

\subsubsection{Observation Modeling}
An observation in our POMDP model is a vector consisting of the vehicle position and the discretized position of all the $n_{ped}$ pedestrians. Given state-of-the-art sensing technology and the effectiveness of filtering techniques, our model assumes no observation noise for these variables (empirically, small noise here does not materially affect agent policy). As a pedestrian's intention is the partially observable variable in our model, we have to infer it from the observations received over time, hedging against estimation uncertainty during decision making.

\subsubsection{Reward Modeling} \label{sec:reward_model}

The POMDP's reward model guides the vehicle towards an optimal driving behavior which is safe, collision-free, and reaches the goal efficiently. We considered the following rewards in our model.
\begin{itemize}
    \item
    Goal Reward: If the vehicle reaches within distance $D_{g}$ to the goal, then there is a large positive reward $R_{goal}$. 
    This reward is modeled to encourage the vehicle to reach its goal.
    \item 
    Obstacle Collision Penalty: If the vehicle passes within a distance $D_{obs}$ to the static obstacle, then there is a substantial negative reward of $R_{obs}$. This reward is modeled to prevent the vehicle from running into static obstacles.
    \item 
    Pedestrian Collision Penalty: If the vehicle is moving and passes within a distance $D_{ped}$ to a pedestrian, then there is a substantial negative reward of $R_{ped}$. If the vehicle is stationary, then we assume the pedestrian is responsible to avoid it. This reward is modeled to ensure safety of the pedestrians as well as the vehicle.
    \item
    Low Speed Penalty: If the vehicle is driving slower than it's maximum possible speed $v_{max}$, then there is a small negative reward $R_{speed} = (v_{c} - v_{max})/v_{max}$. This reward is modeled to encourage the vehicle to drive fast whenever possible.
    \item 
    Sudden Stop Penalty: If the vehicle chooses the $SB$ action, then there is a negative reward of $R_{SB}$. This reward is modeled to incentivize the policy against frequent ``sudden brake" action, and exploring paths where that action can be avoided. 
    \item
    There is also a small negative reward of $R_t$ for every planning step. This reward is included to discourage longer paths. 
\end{itemize}

\subsubsection{Generative Model $G$} \label{sec:gen_function}

For many problems, it is difficult to explicitly represent the probability distributions $T$ and $Z$. Some online POMDP solvers, however, only require samples from the state transitions and observations. As a consequence, it is beneficial to use a generative model which implicitly defines $T$ and $Z$, even when they cannot be explicitly represented. $G$ stochastically generates a new state, observation, and reward given the current state and action: $s', o, r = G(s, a)$. 
In our generative model, for a given POMDP state $s$ and action $a$, we simulate the vehicle forward by applying $a$ for time step $\Delta t$ and move all pedestrians towards their sampled goal location. The $i^{th}$ pedestrian is moved towards $g_i$ by a distance of $v_i\Delta t + \omega_i$, where $\omega_i$ is a small random noise. While more complex pedestrian models exist (e.g. PORCA \cite{luo2018porca}), the choice of dynamic object model is regarded as an interchangeable component of the presented architecture and is not framed as a contribution of this work.

\subsection{Solving POMDPs Online with DESPOT}

% The equations are an exception to the prescribed specifications of this template. You will need to determine whether or not your equation should be typed using either the Times New Roman or the Symbol font (please no other font). To create multileveled equations, it may be necessary to treat the equation as a graphic and insert it into the text after your paper is styled. Number equations consecutively. Equation numbers, within parentheses, are to position flush right, as in (1), using a right tab stop. To make your equations more compact, you may use the solidus ( / ), the exp function, or appropriate exponents. Italicize Roman symbols for quantities and variables, but not Greek symbols. Use a long dash rather than a hyphen for a minus sign. Punctuate equations with commas or periods when they are part of a sentence, as in

% $$
% \alpha + \beta = \chi \eqno{(1)}
% $$

% Note that the equation is centered using a center tab stop. Be sure that the symbols in your equation have been defined before or immediately following the equation. Use Ò(1)Ó, not ÒEq. (1)Ó or Òequation (1)Ó, except at the beginning of a sentence: ÒEquation (1) is . . .Ó

We use a state-of-the-art belief tree search algorithm, DESPOT \cite{ye2017despot} for finding a policy for our POMDP online. Its key strength is handling continuous state space and large observation spaces. To overcome the computational challenge of exploring a large belief tree, DESPOT samples a set of K ``scenarios'', summarizing the execution of all policies under these sampled scenarios. DESPOT builds its tree incrementally by performing a heuristic search guided by a lower bound and an upper bound on the value at each belief node in the tree.  

We calculate the lower bound at a belief leaf node $b_{l}$ by simulating a roll-out policy for all the scenarios at that belief. For $LS$, the roll-out policy executes the hybrid $A^*$ path. For our proposed formulation ($ES$), the roll-out policy executes a path from the vehicle's current location to its goal aided by the use of a multi-query planner (e.g., $PRM$ or $FMM$).  We use a reactive controller to determine vehicle speed along the path. If there are no pedestrians within distance $D_{far}$ from the vehicle, then it increases its speed by 1 $m/s$. If there are pedestrians within distance $D_{near}$ to the vehicle, then it decreases its speed by 1 $m/s$. Otherwise it maintains its current speed. The roll-out policy is run for a fixed, predefined number of steps $M$ or until the termination criteria has been met.

% For figuring out the 2-$D$ action space POMDP planner, the default policy uses the same reactive controller as described above to determine its change in speed. In addition to that, we choose the best steering angle that moves the vehicle towards its goal. The default policy is run for a fixed predefined number of steps $M$. 
% If the vehicle hasn't reached the goal state at the end of the forward rollout, then we give a reward of $R_{goal}/d_{goal}$ at the end where $d_{goal}$ is the euclidean distance between the vehicle and its goal location. This reward is modeled to encourage the vehicle to take the action that takes it closest to its goal.  

We calculate the upper bound at $b_{l}$ by averaging the upper bound for all the scenarios at $b_{l}$. For a scenario, if the vehicle is not stationary and is within distance $D_{ped}$ from any pedestrian, then the bound is $R_{ped}$. Otherwise, it is $\gamma^{t}R_{goal}$ where $t$ is the time taken by the vehicle to reach the goal along the chosen path assuming that the vehicle drives at its maximum speed with no dynamic obstacles (e.g., pedestrians) around. 
% For $LS$, $t$ is the time taken by the vehicle to reach the goal along the hybrid $A^*$ path, assuming that the vehicle drives at its maximum speed with no pedestrians around. For $ES$, $t$ is the time needed needed by the vehicle to reach the goal along the FMM or PRM path under the same assumption.

DESPOT generates a policy tree from this information, with the controller selecting the action at the root of the tree with the greatest expected reward. 

\subsection{Fast Marching Method for Multi-Query Path Planning}  \label{sec:fmm}
The Fast Marching Method ($FMM$) is an algorithm for tracking and modeling the motion of a physical wave interface \cite{osher1988fronts}. The interface is a flat curve in 2-$D$ and a surface in 3-$D$ or higher dimensions. It efficiently solves the Eikonal equation: 
\begin{equation} \label{eikonal_eqn}
1 = F(x) |\nabla T(x)| 
\end{equation}
where x is the position, $F(x) (\geq 0)$ is the expansion speed of the wave at that position, and $T(x)$ is the time taken by the wave interface to reach x from its source.

Given the wave's source point and the expansion speed, $F$ defined over all points in the environment, $FMM$ calculates the time $T$ that the wave takes to reach those points. Since $F>0$, the wave can only expand and it can be shown that the T(x) function (originated by a wave that grows from one single point) has only one global minima at the source and no local minima. 
This method is effective in obtaining a path from any given point in the environment to the wave's source point using gradient descent \cite{fmm_path_planning}. 
% Gomez et al. 2013 \cite{fmm_path_planning} demonstrated $FMM$'s effectiveness in obtaining a path from any given point in the environment to the wave's source point using gradient descent.

% The T(x) function originated by a wave that grows from one single point has only one global minima at the source and no local minima. Since $F>0$, the wave can only expand and the points farther from the source have greater T values. Gomez et al. 2013 \cite{fmm_path_planning} demonstrated how FMM and its variants can be used to obtain a path from any given point in the environment to the wave's source point using gradient descent. 

To find the Eikonal equation solution via FMM, we discretize the environment into grid cells, assigning $F = 0$ for those grid cells where static obstacles are present and $F=1$ everywhere else in the environment. We let the wave originate from the vehicle's goal location and then solve the Eikonal equation using the discrete solution that Sethian proposed in \cite{sethian1999level} to get a grid map of $T$ values for all the cells. We then apply the Sobel operator in a $3\times3$ neighborhood of every grid cell on the grid map to obtain the direction of the gradient at that cell. In order to find a path from any cell in the environment to the vehicle's goal location for the roll-out policy, we move $\alpha$ units in the opposite direction of the gradient at that cell until the goal is reached.

\subsection{Probabilistic Roadmaps for Multi-Query Path Planning}  \label{sec:prm}

The Probabilistic Roadmap ($PRM$) is a well known method for path planning in high dimensions for robots in static environments. The method constructs a graph whose nodes correspond to collision-free configurations in the space and whose edges correspond to feasible paths between these configurations \cite{kavraki1996probabilistic}. This method can be used for any type of holonomic robot. For a holonomic vehicle moving on the 2-$D$ plane, the graph nodes correspond to $(x,y)$ coordinates in the environment and the edges correspond to collision free linear paths between those points. 

In this work, we assigned vehicle's start and goal location as nodes in the $PRM$, and randomly sampled $N_{prm}-2$ more nodes in the environment. We added edges by connecting each node to its $k$ nearest neighbors where the euclidean distance between the two nodes represent the weight of the edge between them. For every node in the $PRM$, we find the shortest path from that node to the vehicle's goal location. In order to find a path from any point in the environment to the vehicle's goal location for the roll-out policy, we first find the nearest $PRM$ node to which straight line traversal is possible and then follow the precomputed path on the $PRM$ from that node to the goal node.   

\vspace{-0.1cm}
\subsection{Tracking POMDP Belief}
The partially observable variables in the POMDP formulation are pedestrian intentions (goal locations) that are inferred by the belief tracker based on the series of observations received. Since in practice there tends to be a finite number of pedestrian goal locations for a given environment, the belief over all such intentions for each pedestrian forms a discrete probability distribution. Changes in goal can also be captured by the POMDP's belief tracker. For each pedestrian being attended to, the belief tracker observes their movement from $(x, y)$ to $(x', y')$, calculates their velocity $v$, and updates the belief $b(g)$ over all the possible intentions to $b'(g)$ using the following update formula: $b'(g) = \eta  p(x', y' | x, y, v, g, M) b(g)$, where $\eta$ is a normalization constant. Based on the chosen pedestrian model $M$, $p(x', y' | x, y, v, g)$ will be directly proportional to the progress the pedestrian made towards goal $g$.

%%%%%%%%%%%%%%%%%%%%%%%%%%%%%%%%%%%%%%%%%%%%%%%%%%%%%%%%%%%%%%%%%%%%%%%%

\section{EXPERIMENTS}

In this section we explain our simulation environment and different experimental scenarios. We also provide specific details about different planners and parameter values used for the experiments. The open source code for the experiments is hosted at \href{https://github.com/himanshugupta1009/extended\_space\_navigation\_pomdp}{https://github.com/himanshugupta1009/extended$\_$space$\_$navigation$\\\_$pomdp}.

%   \begin{figure}[thpb]
%       \centering
%       \framebox{\parbox{3in}{
%       \includegraphics[scale=0.1]{images/scenario_1.png}
%       }}
%     %   \includegraphics[scale=1.0]{images/scenario_1.png}
%       \caption{Inductance of oscillation winding on amorphous
%       magnetic core versus DC bias magnetic field}
%       \label{figurelabel}
%   \end{figure}

\begin{table*}[tb]
\begin{center}
% \begin{tabular}{|p{0.35cm}|p{0.8cm}|p{0.6cm}|p{0.7cm}|p{0.6cm}|p{0.7cm}|p{0.6cm}|p{0.7cm}|}
\caption{Holonomic Vehicle Planner Performance Comparison.}
\label{results_hv}
\begin{tabular}{|p{1cm}|c|c|c|c|c|c|c|c|}
\hline
\multirow{1}{*}{Scenario} & \multicolumn{2}{c|}{1$D$-$A^*$} &
    \multicolumn{3}{c|}{2$D$-$FMM$} & \multicolumn{3}{c|}{2$D$-$PRM$}\\
\cline{2-9}
 (\# Ped) & Time (in s) &  \# SB action & Time (in s) & \# Outperformed & \# SB action & Time  (in s) & \# Outperformed & \# SB action\\
\hline
% \multirow{4}{*}{1} & 100 & 77.02 $\pm$ 0.69 & 0.34 $\pm$ 0.06 & \textbf{64.04 $\pm$ 0.37} & 99 & 0.21 $\pm$ 0.04 & \textbf{64.06 $\pm$ 0.36} & 99 & 0.16 $\pm$ 0.04\\
\cline{1-9} 
 1 (100) & 77.02 $\pm$ 0.69 & 0.34 $\pm$ 0.06 & \textbf{64.04 $\pm$ 0.37} & 99 & \textbf{0.21 $\pm$ 0.04} & \textbf{64.06 $\pm$ 0.36} & 99 & \textbf{0.16 $\pm$ 0.04}\\
\cline{1-9}
 1 (200) & 87.73 $\pm$ 0.86 & \textbf{0.47 $\pm$ 0.06} & \textbf{69.23 $\pm$ 0.54} & 98 & \textbf{0.44 $\pm$ 0.06} & \textbf{69.16 $\pm$ 0.64} & 98 & \textbf{0.36 $\pm$ 0.06}\\
\cline{1-9}
 1 (300) & 99.14 $\pm$ 1.03 & \textbf{0.49 $\pm$ 0.06} & \textbf{76.93 $\pm$ 0.76} & 100 & 0.94 $\pm$ 0.08 & \textbf{76.5 $\pm$ 0.69} & 99 & 0.78 $\pm$ 0.08 \\
\cline{1-9}
 1 (400) & 115.85 $\pm$ 1.81 & \textbf{0.94 $\pm$ 0.10} & \textbf{90.03 $\pm$ 1.42} & 94 & 1.42 $\pm$ 0.12 & \textbf{91.31 $\pm$ 1.09} & 93 & 1.25 $\pm$ 0.11 \\
\hline
% \multirow{4}{*}{2} & 100 & 78.97 $\pm$ 0.86 & 0.31 $\pm$ 0.05 & \textbf{66.05 $\pm$ 0.44} & 95 & 0.24 $\pm$ 0.04 & \textbf{65.92 $\pm$ 0.42} & 93 & 0.16 $\pm$ 0.03 \\
\cline{1-9} 
 2 (100) & 78.97 $\pm$ 0.86 & 0.31 $\pm$ 0.05 & \textbf{66.05 $\pm$ 0.44} & 95 & 0.24 $\pm$ 0.04 & \textbf{65.92 $\pm$ 0.42} & 93 & \textbf{0.16 $\pm$ 0.03} \\
\cline{1-9}
 2 (200) & 89.44 $\pm$ 0.85 & \textbf{0.44 $\pm$ 0.06} & \textbf{72.24 $\pm$ 0.69} & 95 & \textbf{0.48 $\pm$ 0.06} & \textbf{72.21 $\pm$ 0.75} & 98 & \textbf{0.47 $\pm$ 0.06} \\
\cline{1-9}
 2 (300) & 102.67 $\pm$ 1.31 & \textbf{0.5 $\pm$ 0.07} & \textbf{81.95 $\pm$ 0.95} & 98 & 0.94 $\pm$ 0.09 & \textbf{83.74 $\pm$ 1.07} & 93 & 1.05 $\pm$ 0.10 \\
\cline{1-9}
 2 (400) & 114.46 $\pm$ 1.94 & \textbf{0.56 $\pm$ 0.06} & \textbf{90.65 $\pm$ 0.98} & 93 & 1.72 $\pm$ 0.11 & \textbf{92.78 $\pm$ 1.49} & 91 & 1.64 $\pm$ 0.11 \\
\hline
% \multirow{4}{*}{3} & 100 & 82.07 $\pm$ 0.81 & 0.41 $\pm$ 0.06 & \textbf{73.18 $\pm$ 0.64} & 93 & 0.29 $\pm$ 0.04 & \textbf{72.27 $\pm$ 0.66} & 92 & 0.22 $\pm$ 0.04 \\
\cline{1-9}
 3 (100) & 82.07 $\pm$ 0.81 & 0.41 $\pm$ 0.06 & \textbf{73.18 $\pm$ 0.64} & 93 & \textbf{0.29 $\pm$ 0.04} & \textbf{72.27 $\pm$ 0.66} & 92 & \textbf{0.22 $\pm$ 0.04} \\
\cline{1-9}
 3 (200) & 94.87 $\pm$ 1.03 & \textbf{0.35 $\pm$ 0.06} & \textbf{79.53 $\pm$ 0.64} & 93 & 0.56 $\pm$ 0.06 & \textbf{79.23 $\pm$ 0.76} & 91 & 0.49 $\pm$ 0.06 \\
\cline{1-9}
 3 (300) & 107.26 $\pm$ 1.35 & \textbf{0.46 $\pm$ 0.07} & \textbf{87.89 $\pm$ 0.93} & 97 & 0.98 $\pm$ 0.08 & \textbf{87.41 $\pm$ 1.12} & 95 & 1.0 $\pm$ 0.10 \\
\cline{1-9}
 3 (400) & 122.73 $\pm$ 2.32 & \textbf{0.76 $\pm$ 0.09} & \textbf{95.92 $\pm$ 1.02} & 97 & 1.43 $\pm$ 0.11 & \textbf{95.87 $\pm$ 1.20} & 95 & 1.39 $\pm$ 0.10 \\
\hline
% etc. ...
\end{tabular}
\end{center}
\small{The average travel time, the number of trajectories in which each proposed planner outperformed the baseline in terms of travel time, and the average number of $SB$ actions for each algorithm over 100 trials. The standard error of the mean is indicated for averaged quantities. The best travel time and the least amount of $SB$ action in each row are bolded. Multiple entries are bolded if they are statistically similar.}
\end{table*}

\subsection{Simulation Environment}

\begin{figure}
    \centering
    \begin{subfigure}{.155\textwidth}
        \centering
        \includegraphics[width=1\columnwidth]{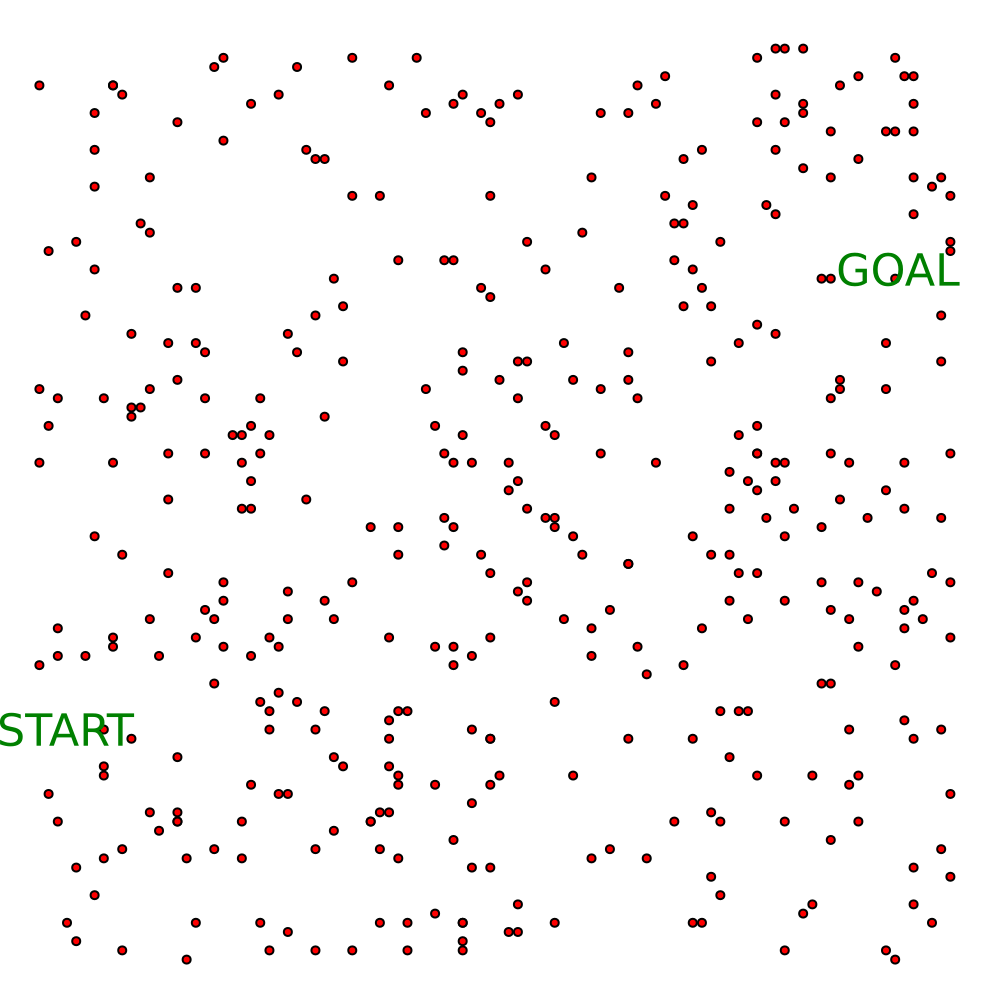}
        \caption{Scenario 1}
        % \caption{Scenario 1: Open field}
        \label{fig:scenario_1}
    \end{subfigure} 
    \begin{subfigure}{.155\textwidth}
        \centering
        \includegraphics[width=1\columnwidth]{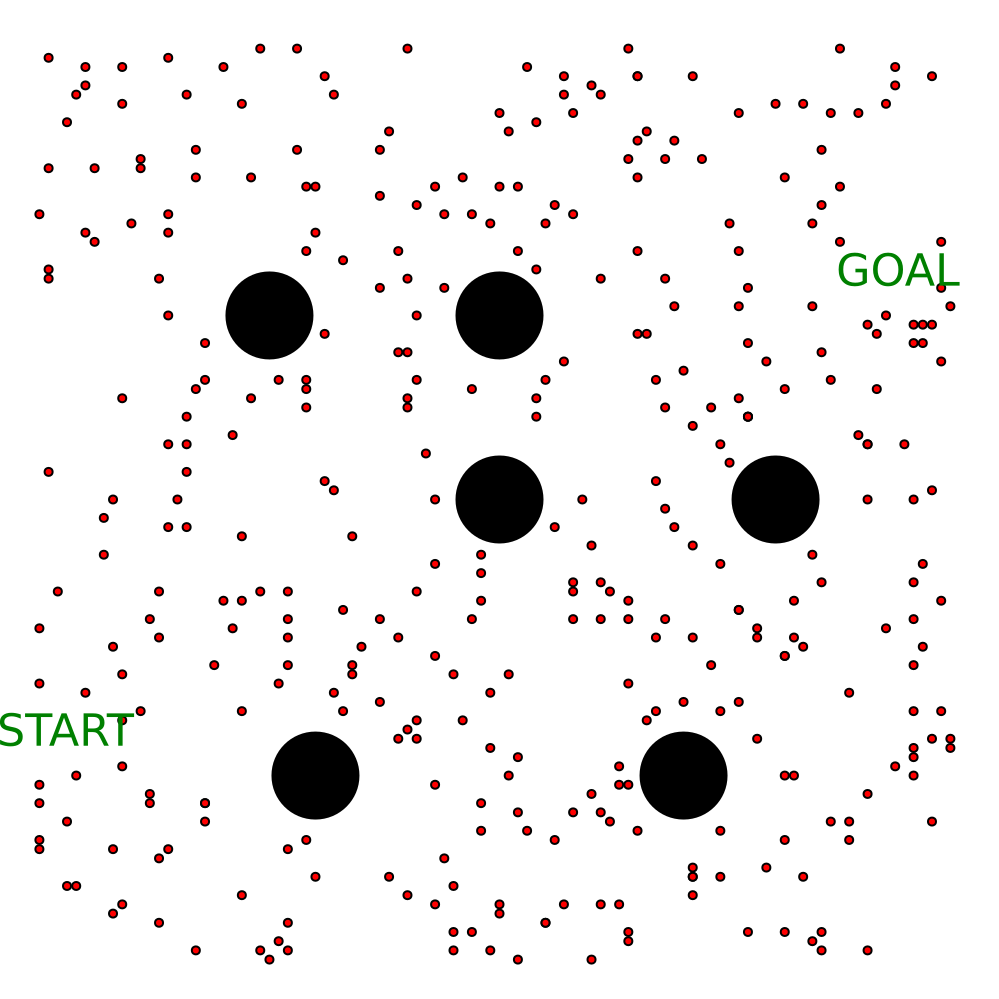}
        \caption{Scenario 2}
        % \caption{Scenario 2: Cafeteria setting}
        \label{fig:scenario_2}
    \end{subfigure}
    \begin{subfigure}{.155\textwidth}
        \centering
        \includegraphics[width=1\columnwidth]{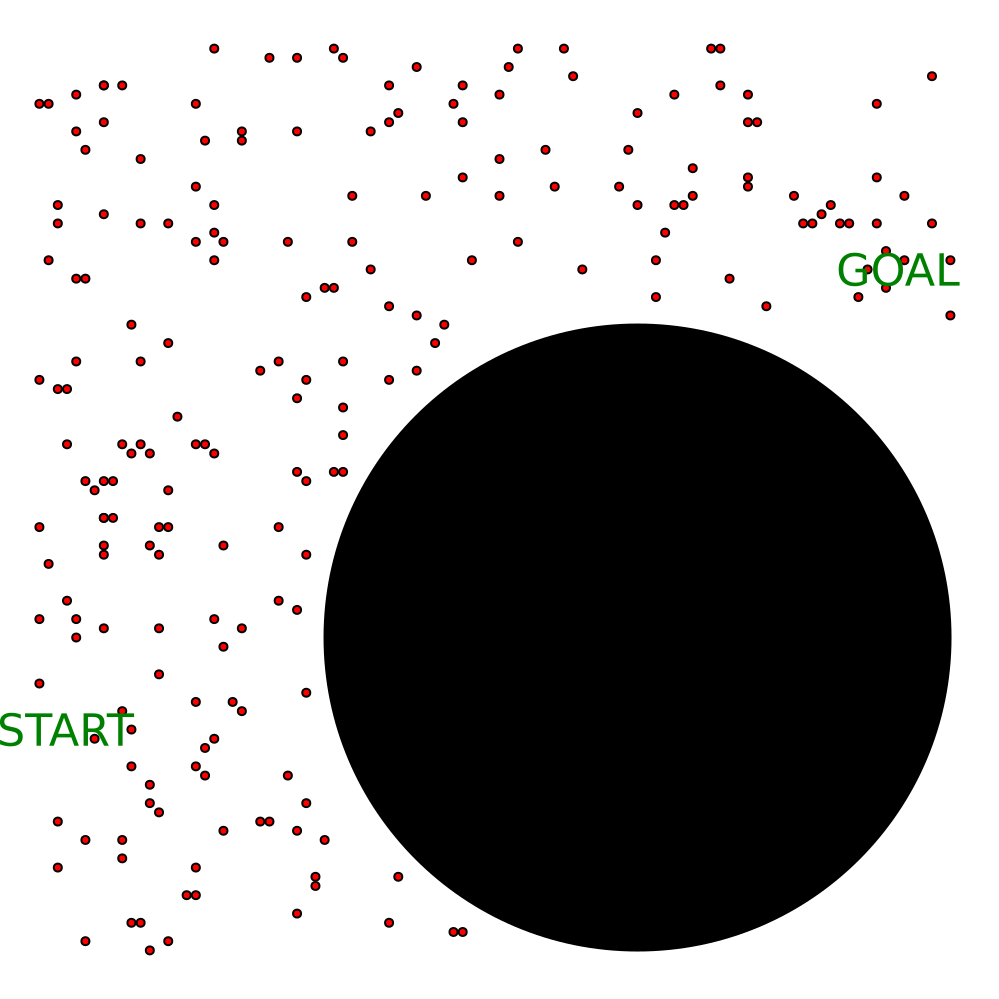}
        \caption{Scenario 3}
        % \caption{Scenario 3: L shaped lobby}
        \label{fig:scenario_3}
    \end{subfigure}
    \caption{Three handcrafted scenarios for evaluating autonomous driving among crowd. The small red dots represent pedestrians and the solid black circles represent static obstacles. In each scenarios, 400 pedestrians are sampled with random initial location and intention. }
    % Scenario 1: No static obstacles; Scenario 2: Multiple small circular obstacles; Scenario 3: Two large circular obstacles.
    \label{fig:environment}

\end{figure}

The environment in our simulator is a $100$ m $\times$ $100$ m square field. The autonomous vehicle is modeled as a holonomic vehicle whose starting position is in the bottom left half and goal location is in the top right half of the field as can be seen in Fig. \ref{fig:environment}. Using the Kinova MOVO robot as a representative example of this class of platform, we set the vehicle's maximum speed to 2 $m/s$.   
Pedestrians are assigned one of the four possible goal locations, located at the corners of the environment. As soon as a pedestrian reaches its goal location, it is removed from the environment and a new pedestrian is spawned randomly along one of the edges of the field. Its goal location is chosen from the two goals on the opposite edge at random. This is done to ensure that there are a fixed number of moving pedestrians in the environment at any moment of time.
The pedestrian simulation model is same as the model used by the POMDP generative function in Section \ref{sec:gen_function}. However, since it is merely a component of our simulator, it can be replaced by an alternative pedestrian model (e.g. PORCA \cite{luo2018porca}) without much effect on the performance of the proposed approach so long as it is congruent with pedestrian behavior in the environment. This simulator was built using the high-level, high-performance programming language Julia \cite{bezanson2017julia}.

\subsection{Experiment Scenarios}
We have designed three different scenarios to compare our planner's performance against a widely adopted baseline method ($LS$ planner) \cite{bai2015intention}. They are as follows:
\subsubsection{Scenario $1$} There are no static obstacles in the environment. It resembles an open field. We designed this to analyze the proposed planner's performance when there is plenty of empty space for the vehicle to explore to avoid collision with pedestrians. 
\subsubsection{Scenario $2$} There are six small static circular obstacles that are scattered throughout the environment. It resembles a cafeteria setting. We designed this to analyze the proposed planner's performance when there is less empty space available but it is distributed throughout the field.
\subsubsection{Scenario $3$} There is a large static circular obstacle in the bottom right corner of the environment. It resembles a L shaped lobby. We designed this to analyze the proposed planner's performance when the empty space is available only in certain parts of the environment which forces the vehicle to navigate among pedestrians in a limited space. 
    
% For each scenario, we ran sets of 100 different experiments for different pedestrian density in the environment. In each experiment, pedestrians were assigned random starting points and intentions, and the number of pedestrians in the environment varied from 100 to 400 (in increments of 100).  

\subsection{Planners}\label{sec:diff_planners}
The alternative experimental planners tested in the experiments are described below.
All planners use the POMDPs.jl~\cite{egorov2017pomdps} implementation of DESPOT from the ARDESPOT.jl package.\footnote{\url{https://github.com/JuliaPOMDP/ARDESPOT.jl}}

\subsubsection{$\textbf{LS}$ \textbf{planner}} \label{sec:a_star}
This is the baseline approach, 1$D$-$A^*$ against which we have compared our proposed planner's performance. At every time step, the hybrid $A^*$ algorithm finds a path from vehicle's current position to its goal. The path generated by Hybrid $A^*$ on this landscape is then used in conjunction with a POMDP solver that determines the optimal speed given the fixed path and pedestrians located around the vehicle.

The Hybrid $A^*$ algorithm is an extension to $A^*$ that can generate a drivable path for the vehicle over continuous state space, and notably was used for autonomous mobile robot path planning during the DARPA Urban Challenge \cite{thrun2006stanley}. 
Hybrid $A^*$ finds a minimum cost path from the vehicle's current position to its goal location. The path cost is the sum of two components, 1) $C_{st}$ to penalize collisions with static obstacles and 2) $C_{ped}$ to penalize collisions with pedestrians. 
We reduce the path cost exponentially over time by a fixed discount factor $\lambda \in (0, 1]$ to more heavily consider cost estimates at the beginning of the path due to increasing uncertainty \cite{bai2015intention}. 
In this approach, every pedestrian is modeled as a static obstacle at the center of a potential field with size proportional to the uncertainty around their intended goal. When pedestrian's intention is highly uncertain, we place a large potential field around the pedestrian's current location. Otherwise, we place a potential field around the pedestrian's most likely path \cite{bai2015intention}.
The path planner has 36 search actions from $-170\degree$ to $180\degree$ at $10\degree$ intervals.

The corresponding POMDP's reward model is the same as that described in Section \ref{sec:reward_model} with the exception of the obstacle collision penalty, which is omitted since collision with static obstacles is avoided by the path planner. $LS$ uses DESPOT to determine the best possible $\delta_s$ out of $\{-1 m/s, 0 m/s, 1 m/s, SB\}$ at every time step along the generated path.
    
\subsubsection{$\textbf{ES}$ \textbf{planner}}
We propose two $ES$ planners, $2D$-$FMM$ and $2D$-$PRM$. At every time step, $ES$ selects both, $\delta_{\theta}$ and $\delta_{s}$ for the vehicle. Since DESPOT does not perform well for continuous or large action space problems, we choose a small discrete set of actions depending on the vehicle's state variables $(x_c,y_c,v_c)$. When $v_c = 0$, there are 9 possible actions. It can either stay stationary (i.e $\delta_{s}$ = 0 $m/s$) or increase its speed (i.e $\delta_{s}$ = 1 $m/s$) while choosing a $\delta_{\theta}$. There are 7 possible choices for $\delta_{\theta}$ from  $-45\degree$ to $45\degree$ at $15\degree$ intervals. We add another potential value for $\delta_{\theta}$ related to potential roll-out policies called $\delta_{RO}$, which changes the vehicle's orientation according to the $FMM$ or $PRM$ roll-out policy at $(x_c,y_c)$.
If $v_c \neq 0$, then there are 11 possible actions. The planner can choose to either increase ($\delta_{s}$ = 1 $m/s$) or decrease ($\delta_{s}$ = -1 $m/s$) its speed without changing its orientation (i.e. $\delta_{\theta} = 0\degree$), or maintain its current speed (i.e $\delta_{s}$ = 0 $m/s$) and select from 8 possible $\delta_{\theta}$ choices mentioned above, or apply the SB action. 

Depending on the planner, DESPOT uses either $FMM$ (Section \ref{sec:fmm}) or $PRM$ (Section \ref{sec:prm}) to obtain a path for the scenarios at a belief node. To evaluate the lower bound at that belief node, the roll-out policy executes a reactive controller to determine speed over that path.
% DESPOT calculates a lower bound for every belief node during tree search by executing a roll-out policy for the scenarios at that belief. 
% For every state encountered during the tree search, DESPOT uses either $FMM$ or $PRM$ planning technique depending on the planner to get a path from that state to the goal. It then executes a roll-out policy with a reactive controller on that path to evaluate the lower bound for the belief node during tree search. 

\begin{table*}[tb]
\caption{Non-Holonomic Vehicle Planner Performance Comparison.}
% \caption{COMPARISON OF PERFORMANCE OF DIFFERENT PLANNERS FOR A NON-HOLONOMIC VEHICLE.}
\begin{center}
% \begin{tabular}{|p{1.5cm}|p{1.4cm}|p{1.4cm}|p{1.4cm}|p{1.35cm}|p{1.35cm}|}
\begin{tabular}{|c|c|c|c|c|c|}
\hline
\multirow{1}{*}{} & \multicolumn{2}{c|}{1$D$-$A^*$} & \multicolumn{3}{c|}{2$D$-$NHV$}\\
\cline{1-6}
\#Ped & Time (in s) & \# SB action  &  Time (in s) & \# Outperformed & \# SB action\\
\hline
100 & 53.62 $\pm$ 0.91 & 1.84 $\pm$ 0.11 & \textbf{36.54 $\pm$ 0.45} & 98 & \textbf{0.26 $\pm$ 0.04}\\
\hline
200 & 72.54 $\pm$ 1.09 & 2.84 $\pm$ 0.14 & \textbf{43.79 $\pm$ 0.85} & 96 & \textbf{1.04 $\pm$ 0.09}\\
\hline
300 & 95.43 $\pm$ 1.74 & 3.62 $\pm$ 0.15 & \textbf{62.37 $\pm$ 1.57} & 95 & \textbf{2.55 $\pm$ 0.12}\\
\hline
400 & 110.41 $\pm$ 2.32 & 3.98 $\pm$ 0.15 & \textbf{81.07 $\pm$ 1.80} & 95 & \textbf{3.54 $\pm$ 0.13}\\
\hline
\end{tabular}
\end{center}
\label{results_nhv}
% \small{For scenario 1, we ran 100 experiments for different number of pedestrians in the environment. The average travel time, the number of trajectories in which the proposed planner outperformed the baseline in terms of travel time, and the average number of $SB$ actions is calculated using all the trials.} 
\small{This table shows the average travel time, the number of trajectories in which the proposed planner outperformed the baseline in terms of travel time, and the average number of $SB$ actions over 100 trials under Scenario 1. The standard error of the mean is indicated for averaged quantities. The best travel time and the least amount of $SB$ action in each row are bolded.}
\vspace{-4mm}
\end{table*}

\begin{figure*}
    \centering
    
    \begin{subfigure}{.2\textwidth}
        \centering
        \includegraphics[width=1\columnwidth]{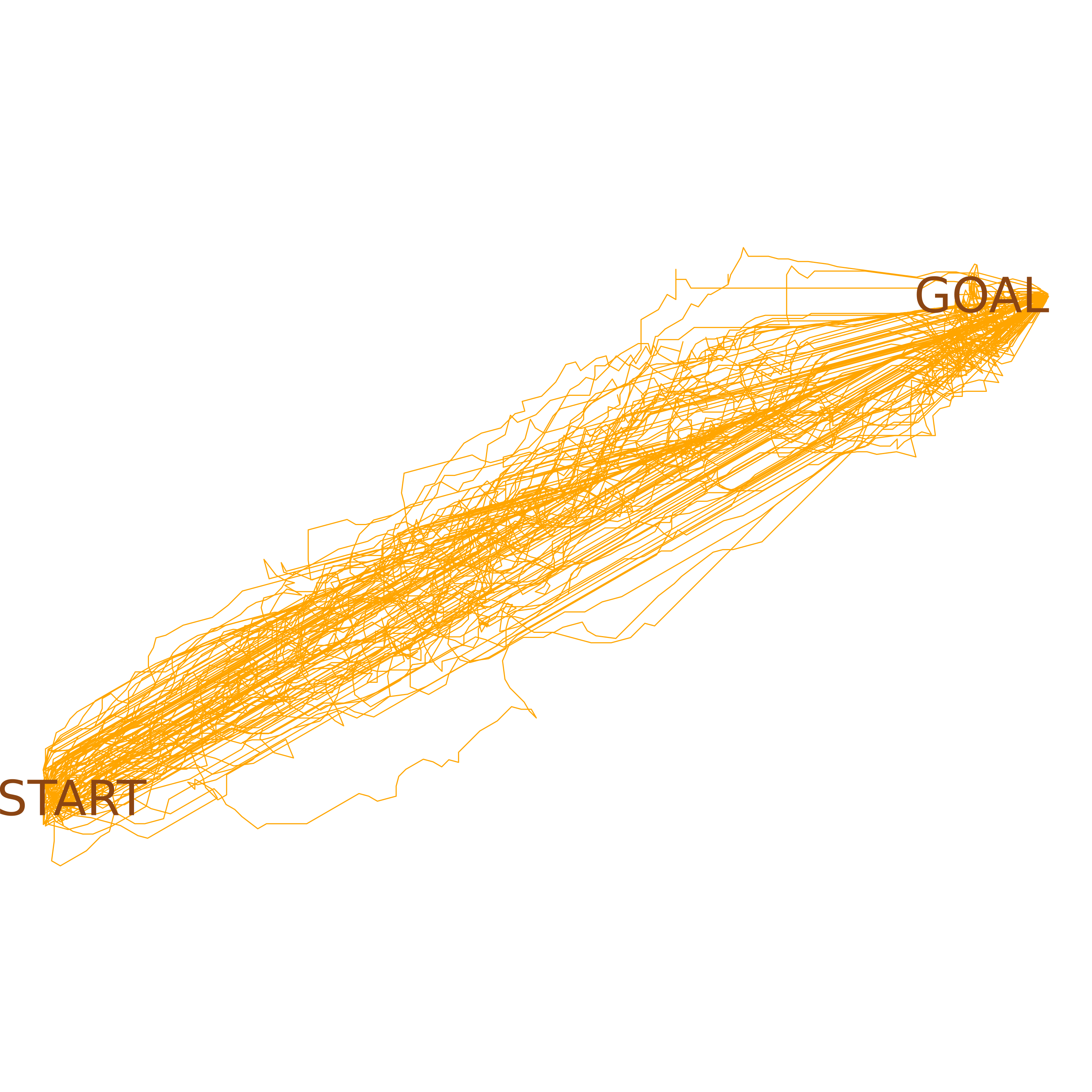}
        \caption{1D-A*}
        \label{fig:traj_sce1_astar}
    \end{subfigure} 
    \begin{subfigure}{.2\textwidth}
        \centering
        \includegraphics[width=1\columnwidth]{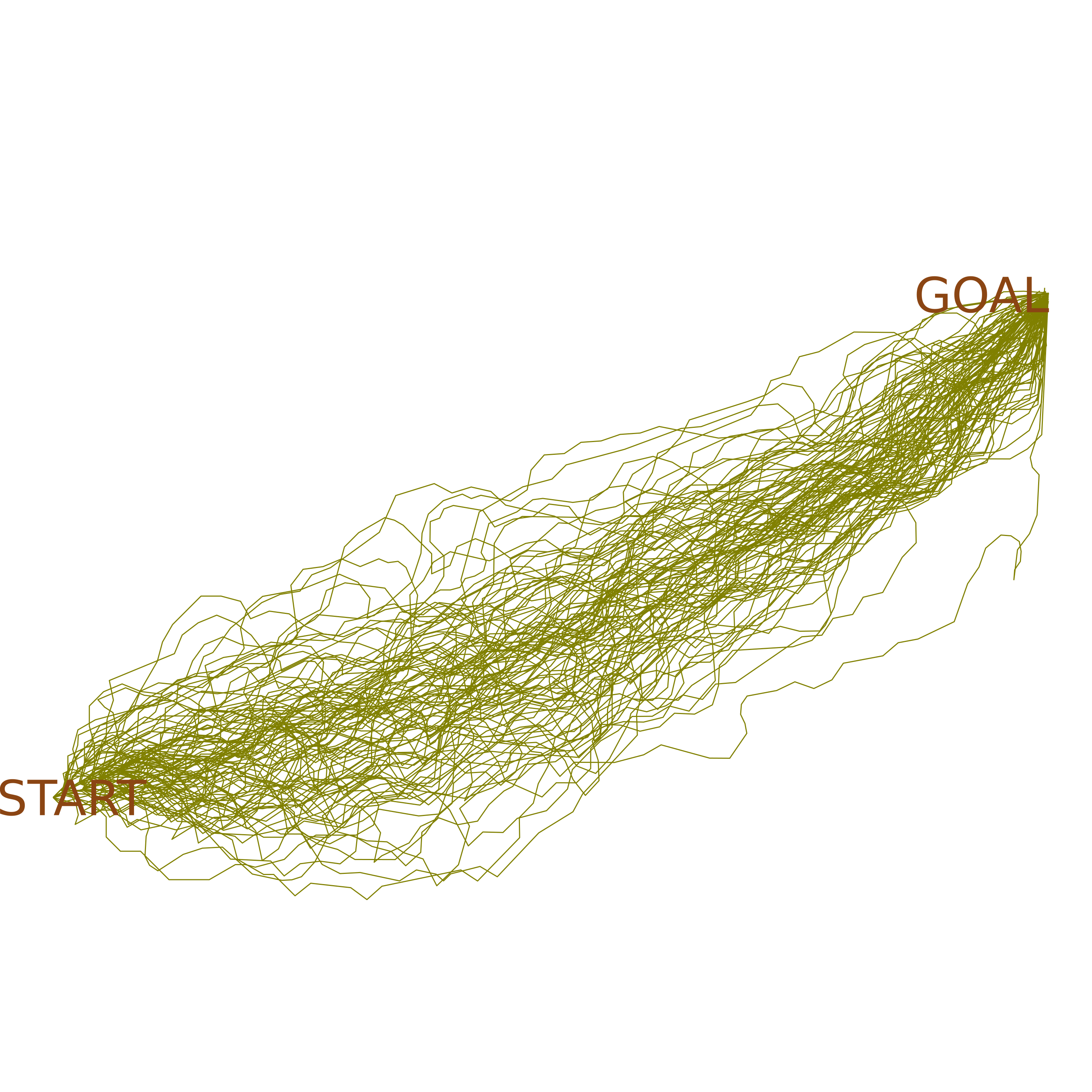}
        \caption{2D-FMM}
        \label{fig:traj_sce1_fmm}
    \end{subfigure}
    \begin{subfigure}{.2\textwidth}
        \centering
        \includegraphics[width=1\columnwidth]{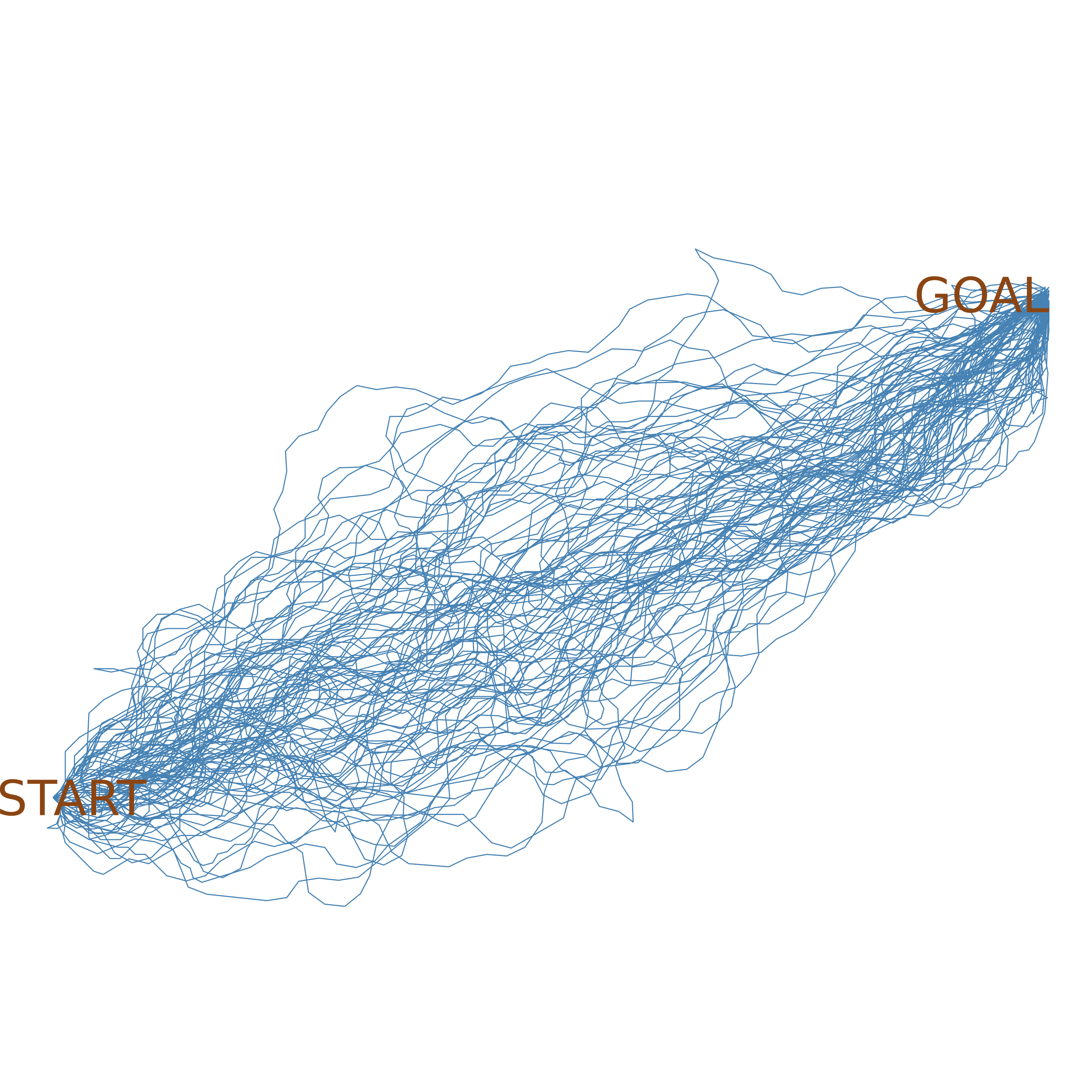}
        \caption{2D-PRM}
        \label{fig:traj_sce1_prm}
    \end{subfigure}

    % \caption{Trajectories executed by the holonomic vehicle using different planners across all 100 experiments in Scenario 1.}
    % \label{fig:traj_comparison}

    \begin{subfigure}{.2\textwidth}
        \centering
        \includegraphics[width=1\columnwidth]{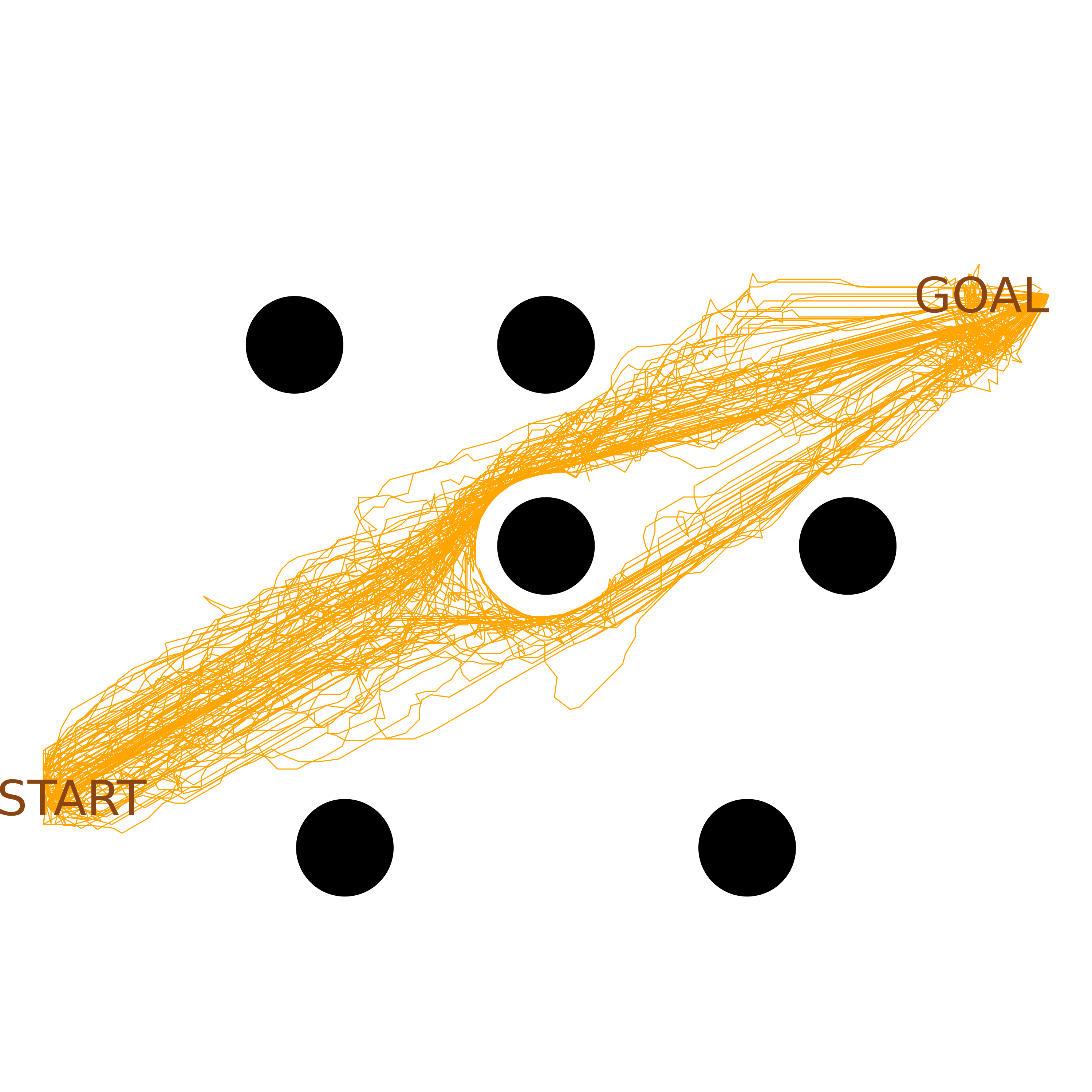}
        \caption{1D-A*}
        \label{fig:traj_sce2_astar}
    \end{subfigure} 
    \begin{subfigure}{.2\textwidth}
        \centering
        \includegraphics[width=1\columnwidth]{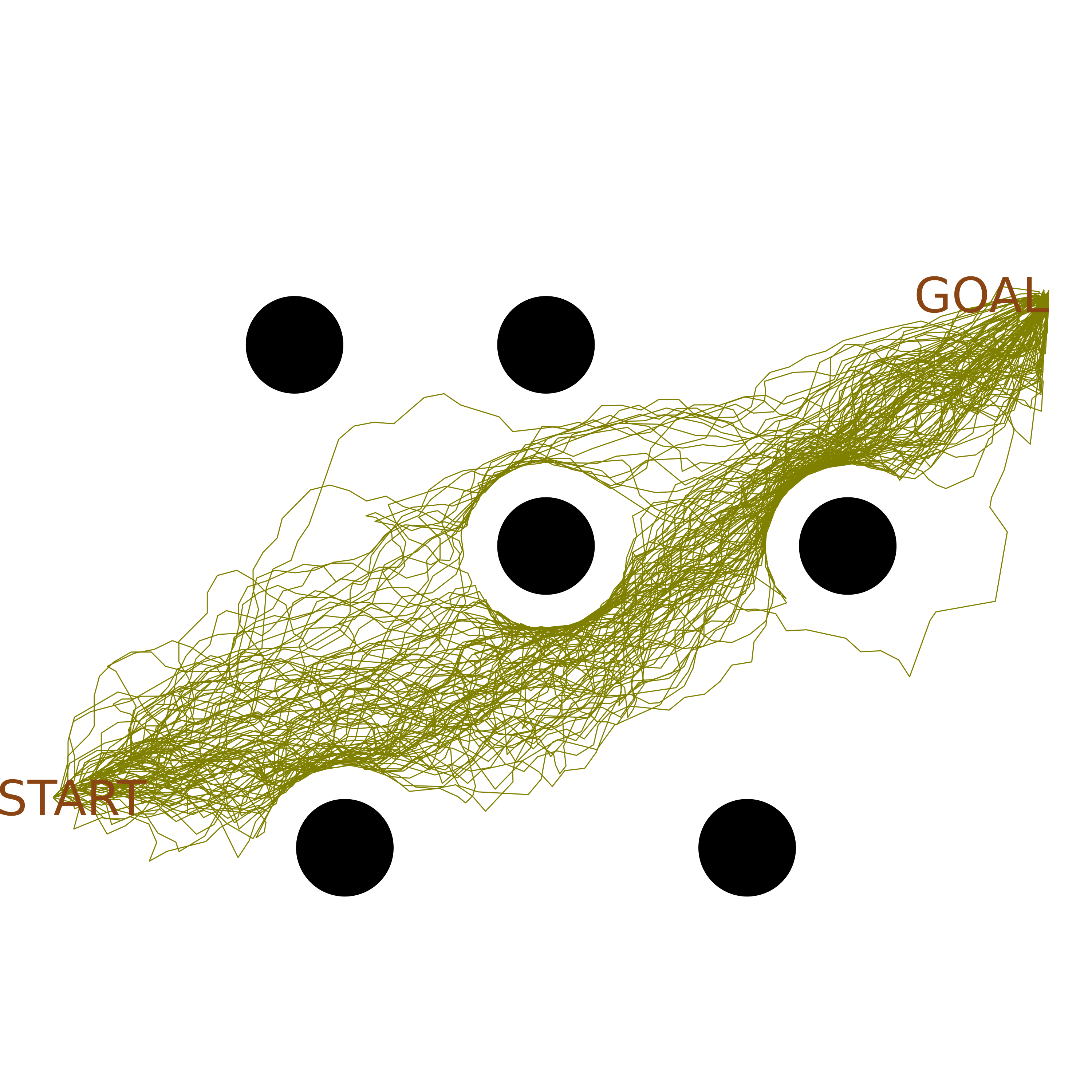}
        \caption{2D-FMM}
        \label{fig:traj_sce2_fmm}
    \end{subfigure}
    \begin{subfigure}{.2\textwidth}
        \centering
        \includegraphics[width=1\columnwidth]{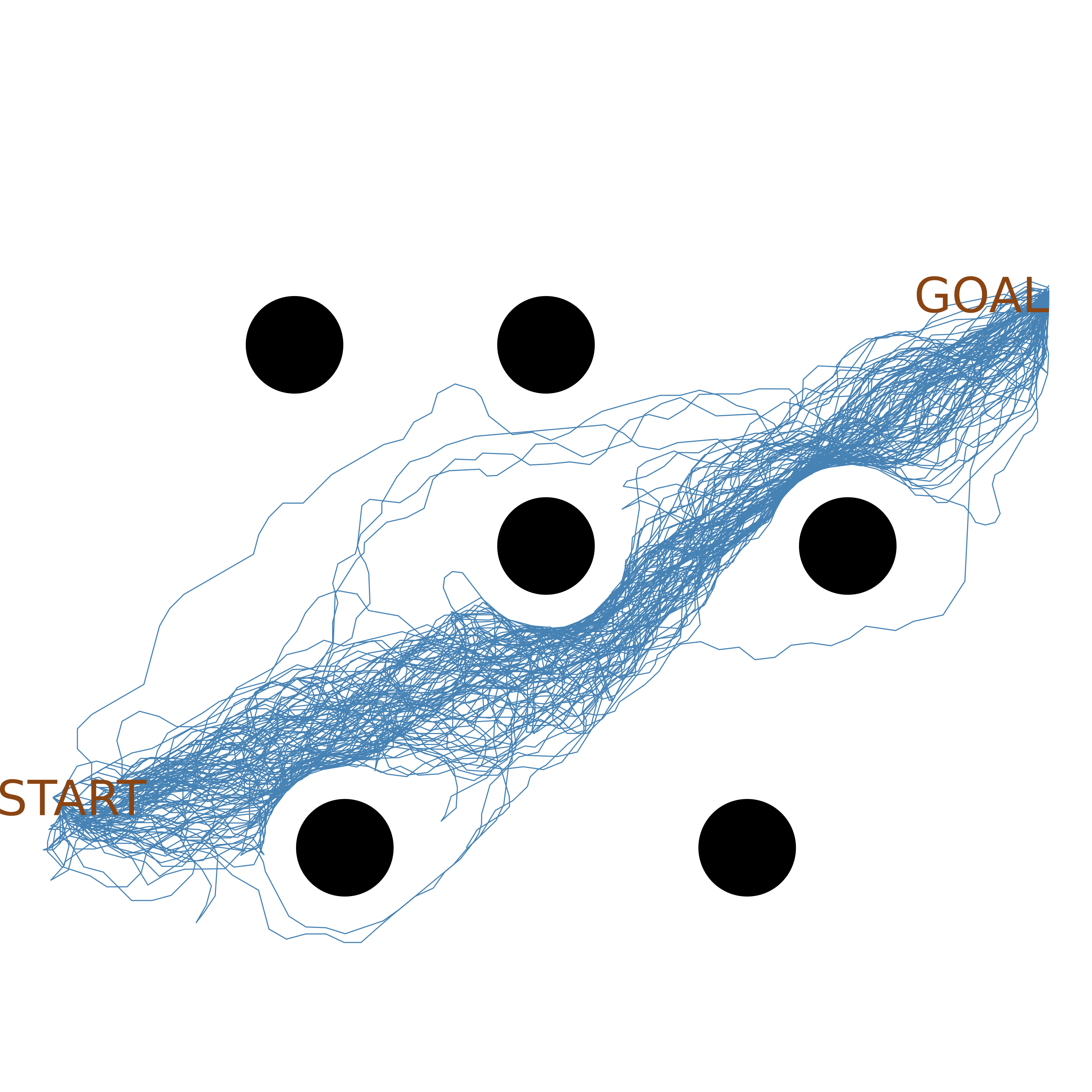}
        \caption{2D-PRM}
        \label{fig:traj_sce2_prm}
    \end{subfigure}

    \begin{subfigure}{.19\textwidth}
        \centering
        \includegraphics[width=1\columnwidth]{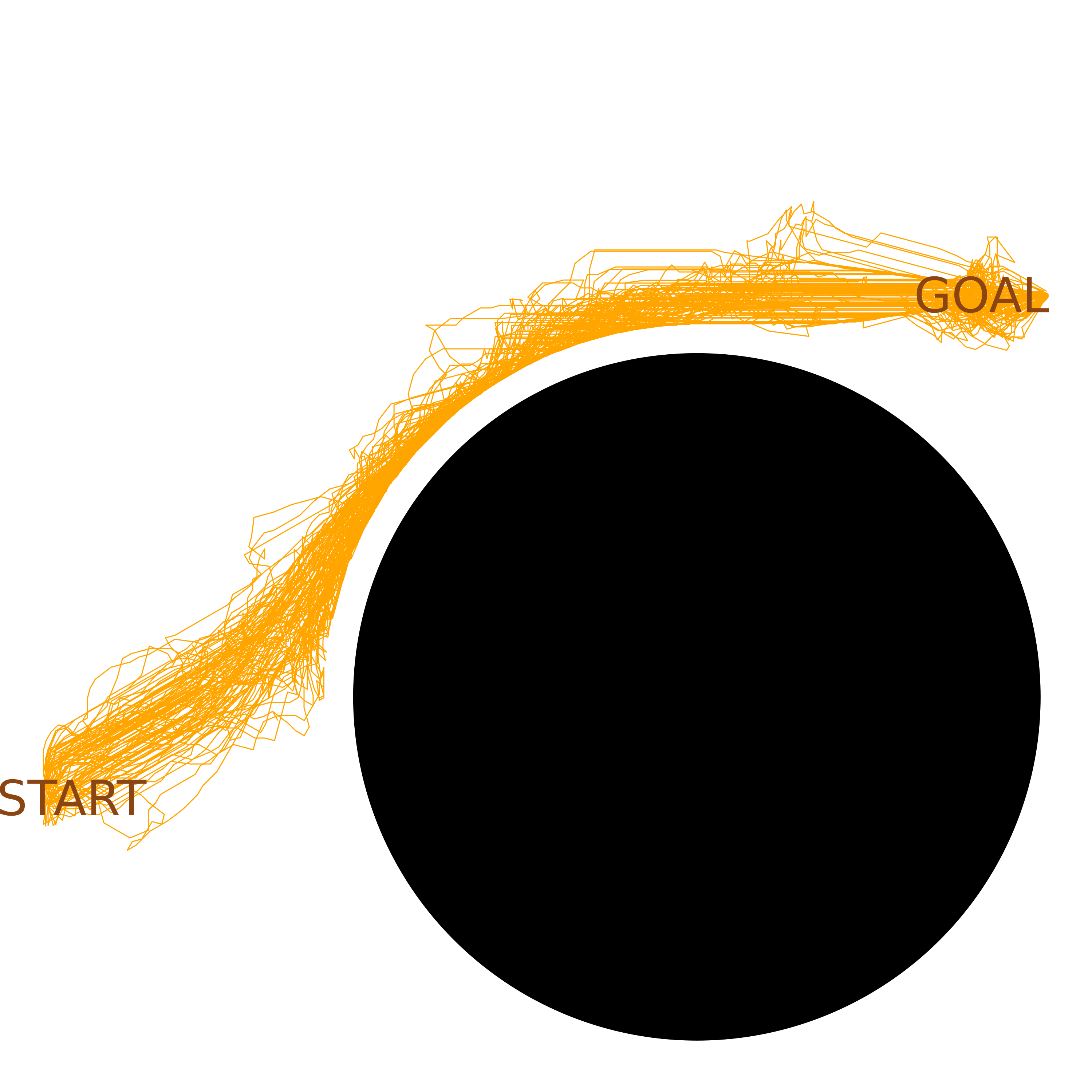}
        \caption{1D-A*}
        \label{fig:traj_sce3_astar}
    \end{subfigure} 
    \begin{subfigure}{.19\textwidth}
        \centering
        \includegraphics[width=1\columnwidth]{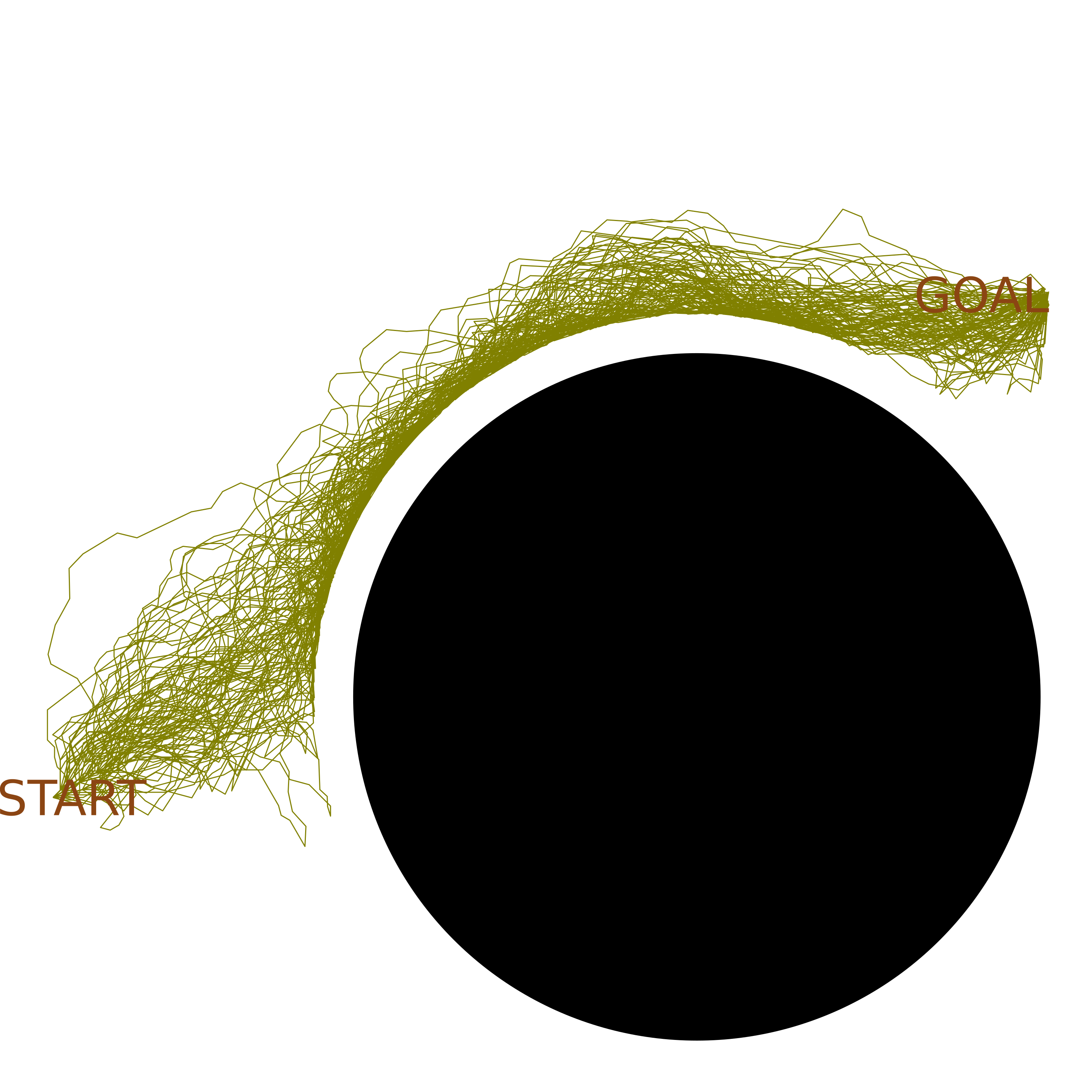}
        \caption{2D-FMM}
        \label{fig:traj_sce3_fmm}
    \end{subfigure}
    \begin{subfigure}{.19\textwidth}
        \centering
        \includegraphics[width=1\columnwidth]{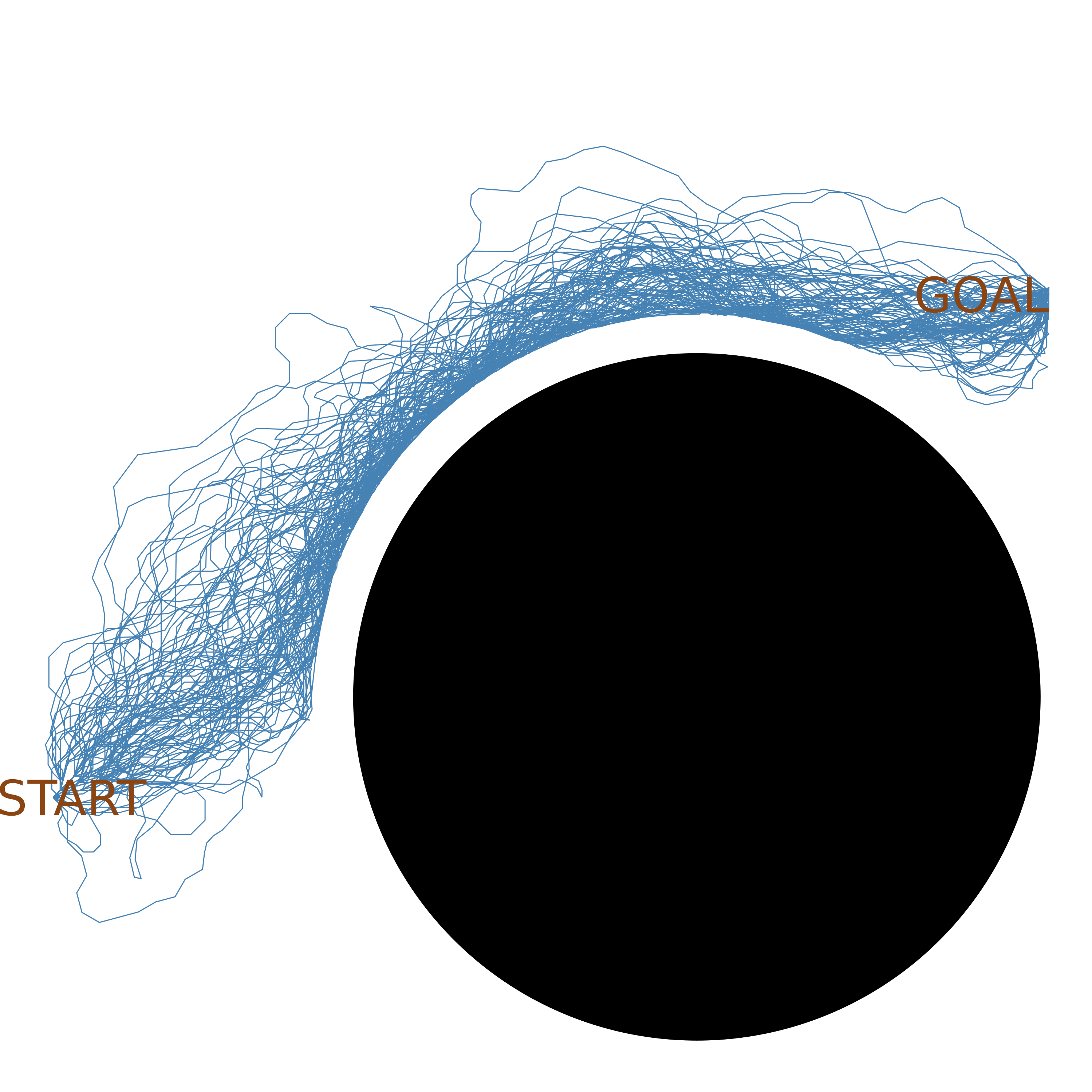}
        \caption{2D-PRM}
        \label{fig:traj_sce3_prm}
    \end{subfigure}

    \caption{Trajectories executed by the holonomic vehicle using different planners across all 100 experiments in Scenario 1 (Fig. \ref{fig:traj_sce1_astar}, \ref{fig:traj_sce1_fmm}, \ref{fig:traj_sce1_prm}), Scenario 2 (Fig. \ref{fig:traj_sce2_astar}, \ref{fig:traj_sce2_fmm}, \ref{fig:traj_sce2_prm}), and Scenario 3 (Fig. \ref{fig:traj_sce3_astar}, \ref{fig:traj_sce3_fmm}, \ref{fig:traj_sce3_prm}) with 400 pedestrians in the environment.}
    \label{fig:traj_comparison}
    \vspace{-4mm}
\end{figure*}

\subsection{Experimental Details}
For each scenario, we ran sets of 100 different experiments with different pedestrian density in the environment. The number of pedestrians in the environment varied from 100 to 400 (in increments of 100). In each experiment, pedestrians were assigned random starting points and intentions. The performance of different planners was compared for that sampled environment under the same random seed for noise in simulated pedestrian motion. In simulations, the planning time for each step is 0.5 seconds. For 1$D$-$A^*$, we devote 0.15 seconds for path planning, and 0.35 seconds for speed planning by solving the corresponding POMDP. For 2$D$-$FMM$ and 2$D$-$PRM$, all of the planning time is devoted to solving the POMDP because the multi-query motion planning needs to be computed only once for the environment. The online POMDP solver reasons over the uncertainty in intentions of 6 nearest pedestrians (i.e. $n_{ped} = 6$). DESPOT performs online tree search with 100 sampled scenarios. We define a trajectory to be unsafe if at any time step the moving vehicle gets within $1$ m distance of a pedestrian. 
% An experiment is considered successful if the vehicle reaches its goal within a time limit of 300 seconds. 

% In that case an effective roll-out policy is to change the vehicle's orientation to move in a straight line path towards the goal but restricting it planner. This can easily be extended to scenarios with static obstacles by using the Fast Marching Square method for non holonomic planning as described in \cite{arismendi2015nonholonomic} for the roll-out policy. to the  stay all steering planners have the angle limit. The their-$D$ actionthey executed  safe trajectoriesconstantly them, and the planner outperformed always theDP 

%%%%%%%%%%%%%%%%%%%%%%%%%%%%%%%%%%%%%%%%%%%%%%%%%%%%%%%%%%%%%%%%%%%%%%%%

\section{RESULTS AND DISCUSSION}

\begin{figure}
    \centering
        \includegraphics[width=.8\columnwidth]{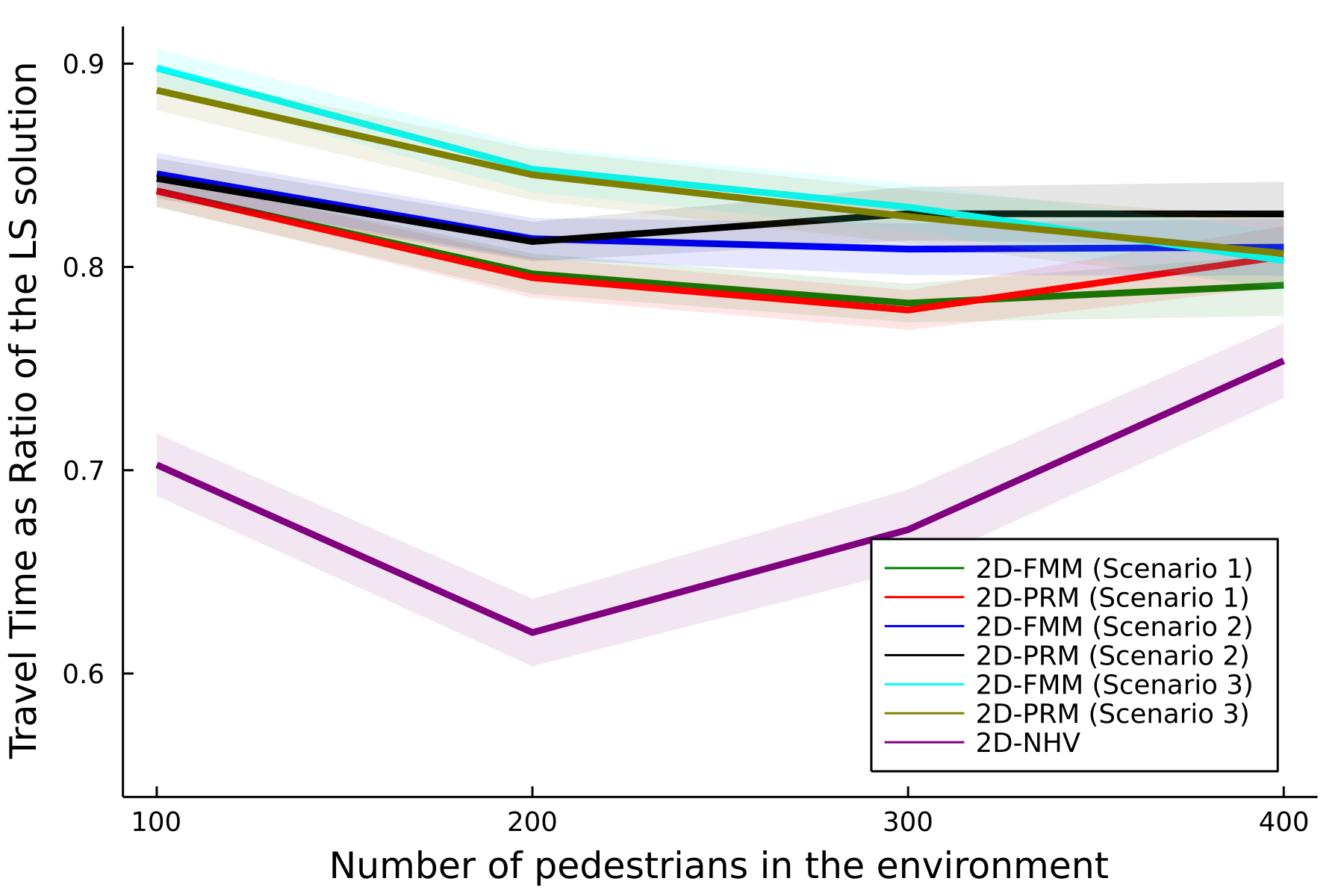}
        \caption{ES planners consistently outperform LS planners across each scenario, whether for holonomic or non-holonomic agents.}
        \label{fig:time_results}
        \vspace{-6mm}
\end{figure}

The results from our experiments are summarized in Tables \ref{results_hv} and \ref{results_nhv}.
We computed the average travel time, the number of trajectories in which the proposed planner outperformed the baseline in terms of travel time, and the average number of times $SB$ action was executed across 100 experiments for all the planners in different settings. 
Since all the planners have the $SB$ action in their action space, they executed safe trajectories in every experiment. All of the experiments fulfilled the success criteria.

Experimental results for a holonomic vehicle are compiled in Table \ref{results_hv}. 
The first key observation is that our proposed extended space planners, 2$D$-$FMM$ and 2$D$-$PRM$ executed paths that took less travel time than the baseline without compromising safety. For each experimental setting, the best travel time is marked in bold in Table \ref{results_hv}. The travel time for 2$D$-$FMM$ and 2$D$-$PRM$ is comparable across all the different settings which indicates that the sensitivity of travel time to the choice of motion planning algorithm used for generating effective roll-out policies is low. The differences in travel time for $ES$ planners as compared to the $LS$ planner across different settings are characterized in Fig. \ref{fig:time_results}.

The baseline approach took more travel time on average primarily due to the segregation of planning problem in two components. 
The hybrid $A^*$ algorithm generates a path without considering the vehicle's speed and by using ad-hoc techniques to handle uncertainty in pedestrian intention as described in Section \ref{sec:a_star}.
In most of the cases, the POMDP speed planner ($LS$) realizes that traveling along this fixed path at the vehicle's current speed can lead to a collision. As a result, it decides to either slow down or stop which increases the travel time.
The decoupling of heading angle and speed forces the baseline approach to reason over uncertainty along just one path and the vehicle fails to perform efficient motion between moving pedestrians. 
On the other hand, the $ES$ planners reason over uncertainty along multiple paths (Fig. \ref{explored_trajectories}) and often manage to find a path where they do not have to slow down or stop.
Moreover, it is possible that the hybrid $A^*$ path might not be obtained at every time step within the limited computation time. As a result, the system has to estimate the speed over the old path which was constructed considering the position of pedestrians and belief over their intention at the previous time step. 
This leads to sub-optimal decision making.

% By isolating the planning problem into two components, the baseline approach lost the capability to effectively plan through time. Due to decoupling of heading angle and speed, it reasons over uncertainty along just one path and the vehicle fails to perform narrow collision free motion between moving pedestrians. On the other hand, the $ES$ planners reason over uncertainty along multiple paths to the goal as shown in Fig. \ref{explored_trajectories}.
% When surrounded by pedestrians, the baseline approach decides to just stay on the hybrid $A^*$ path instead of finding narrow collision free paths between moving pedestrians. 
% For instance, assume the case where the belief tracker is fairly certain about a pedestrian's intention that is in the vehicle's vicinity. In 1$D$-$A^*$, the path planner plans through time and generates a path by treating the pedestrian as a static obstacle at its most likely position in the future. Since the path is generated assuming a fixed vehicle speed, it's possible that the POMDP speed planner realizes  vehicle's current speed is different speed planner  

Another important observation is that both proposed planners outperformed the baseline approach in the metric of travel time at least 91\% of the times across all the different settings. 
In densely crowded environments, moving to empty spaces nearer to the agent's goal (instead of staying idle and letting pedestrians pass) is an intuitively good strategy. This behavior is visible from the trajectories executed by 2$D$-$PRM$ and 2$D$-$FMM$ (Fig.\ref{fig:traj_comparison}). They cover a wider area of the environment than  1$D$-$A^*$.   
However, doing so can sometimes also result in the vehicle momentarily getting stuck behind a group of pedestrians that it did not reason over earlier. This happened in the few experiments where the proposed planners took more travel time. 
% This can be improved by using better cost models  
% \hg{Should we talk about how this can be improved if we use better cost models for motion planning techniques? I have currently put that in Future Work section, but I think we should mention that here.}

The extended space planners executed more $SB$ actions on average than the baseline for settings with high pedestrian density under all the scenarios. 
This is mainly due to the availability of more choices of $\delta_{\theta}$ for $LS$ (36 choices) than $ES$ (8 choices). 
The result with the least amount of $SB$ action is marked in bold for every setting in Table \ref{results_hv}.
For low pedestrian density, $ES$ planners took less or almost the same amount of $SB$ actions as the $LS$ planner. 
Densely populated environments have less free space for the vehicle to navigate. In that situation, having more choices for $\delta_{\theta}$ allows the $LS$ planner to find a path to move to open spaces that avoids collision with any pedestrian. 
On the contrary, due to limited choices, the $ES$ planner decides to execute the $SB$ action under the same situation. 
This is not a limitation directly due to extending the action space; rather it is a limitation of the particular online tree search algorithm, DESPOT, which does not work well for large or continuous action space problems. 
This demonstrates a need for online POMDP solvers that can better solve continuous action space problems which is an active area of research \cite{lim2021voronoi}.

We also performed experiments for a non-holonomic vehicle and compared the performance of $LS$ and $ES$ planners under Scenario 1. The vehicle is modeled as a Dubin's car with a max speed of 4 $m/s$. For $LS$, the hybrid $A^*$ path planner has 19 search actions from $-45\degree$ to $45\degree$ at $5\degree$ intervals, and the POMDP model is same as that for a holonomic vehicle. For $ES$, an effective roll-out policy in the absence of static obstacles is to apply a steering angle $\beta$ that modifies the vehicle's heading angle to follow a straight line path to the goal (subject to steering angle constraints). $\delta_{RO}$  can be calculated from $\beta$ using the vehicle dynamics. The straight line roll-out is not effective in scenarios 2 and 3 as it could lead to collisions with static obstacles. Under those scenarios, a possible effective roll-out policy for a non-holonomic vehicle would be to execute a reactive controller over the path generated using the Fast Marching Square method \cite{arismendi2015nonholonomic}. 

\begin{figure}
    \centering
    \begin{subfigure}{.23\textwidth}
        \centering
        \includegraphics[width=1\columnwidth]{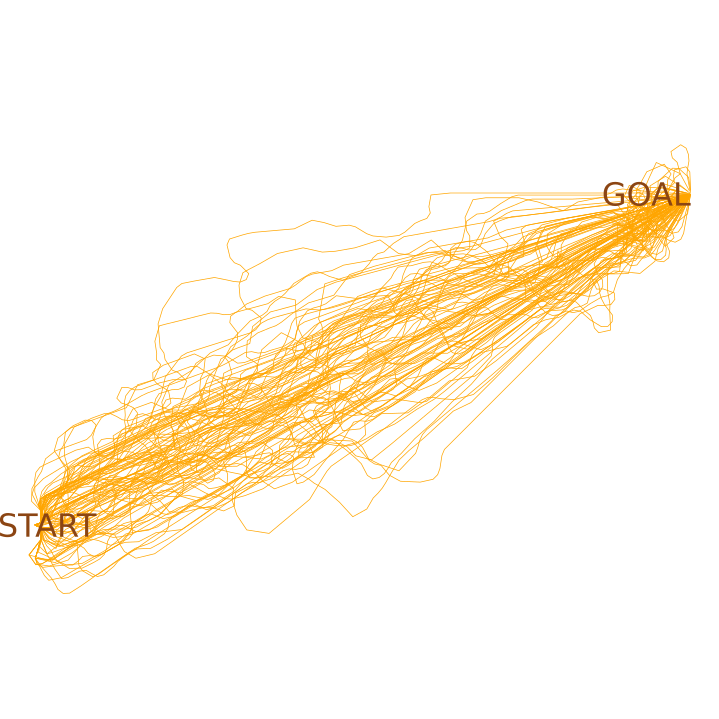}
        \caption{1D-A*}
        \label{fig:nhv_traj_astar}
    \end{subfigure} 
    \begin{subfigure}{.23\textwidth}
        \centering
        \includegraphics[width=1\columnwidth]{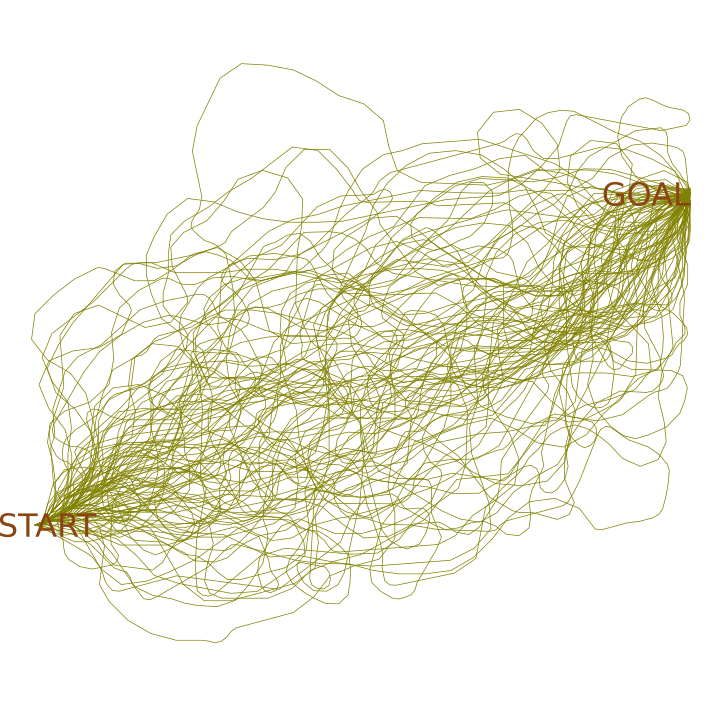}
        \caption{2D-NHV}
        \label{fig:nhv_traj_sl}
    \end{subfigure}
    
    \caption{Trajectories executed by the non-holonomic vehicle using different planners across all 100 experiments in Scenario 1 with 400 pedestrians in the environment.}
    % Scenario 1: No static obstacles; Scenario 2: Multiple small circular obstacles; Scenario 3: Two large circular obstacles.
    \label{fig:nhv_traj}
    \vspace{-4mm}
\end{figure}

The experimental results with the non-holonomic vehicle are summarized in Table \ref{results_nhv}. 
The $ES$ planner (2$D$-$NHV$) took less travel time on average than the $LS$ planner (1$D$-$A^*$) across each of the different settings, outperforming the baseline approach in this metric in at least 95\% of the simulations. 
The sub-optimality in decision making due to the decoupling of heading and speed becomes more significant when the vehicle can travel at higher speeds. 
This is visible from the larger reduction in the travel time ratio for the non-holonomic vehicle (max speed = 4 $m/s$) in comparison to the holonomic vehicle (max speed = 2 $m/s$) across the different pedestrian densities (Fig. \ref{fig:time_results}). 
Since the number of choices of $\delta_{\theta}$ for $LS$ planner (19 choices) is not significantly more than the number of choices for the $ES$ planner (8 choices) for the non-holonomic vehicle, and due to decoupling, the $LS$ planner executes more $SB$ actions than the $ES$ planner on average across all settings. The $ES$ planner plans effectively and finds paths by covering a much wider space of the environment without having to suddenly stop (Fig. \ref{fig:nhv_traj}).

\section{CONCLUSION} \label{conclusion}
% Bullet points : 
% 1) Our approach is better. So, people  should use this from now on!
% 2) FM2 can be used for non holonomic vehicle
% 3) PRM can be used for high dimensional problems
% 4) This shows the need for continuous action space POMDP solvers, and that is something suthors are interested in solving.
% 5) Extend the work to incorporate goal object association and then modify path costs for PRM and FMM.
% 6) Apply this approach for manipulators and on an actual hardware.

This work presents an intention-aware navigation system that uses an extended-space POMDP planner to generate efficient navigation policies in the presence of uncertainty introduced by other agents in the environment.
% demonstrated that for the problem of autonomous navigation in crowded environments solving extended space POMDP that has control over all relevant degrees of freedom, instead of a select few results in significantly efficient trajectories. 
In particular, for an autonomous vehicle navigating among pedestrians, solving a POMDP that controls all degrees of freedom(i.e. speed and heading), instead of a select few (i.e. speed over the Hybrid $A^*$ path) results in faster trajectories. 
This additional control parameter enlarges the reachable state space and raises the need for practical roll-out policies from these states to find the sparse positive reward during the online tree search. Our system uses multi-query motion planning techniques like Fast marching Methods or Probabilistic Roadmaps to efficiently generate effective roll-out policies. 
Our results show that the proposed extended space POMDP planners enable effective and safe autonomous driving in complex crowded environments. They also indicated that the choice of the motion planning technique for offloading the task of generating effective roll-out policies does not affect the planner's performance significantly. 

\begin{acks}
This work was funded in part by NSF NRI Award \#1830686.
\end{acks}

%%%%%%%%%%%%%%%%%%%%%%%%%%%%%%%%%%%%%%%%%%%%%%%%%%%%%%%%%%%%%%%%%%%%%%%%

%%% The next two lines define, first, the bibliography style to be 
%%% applied, and, second, the bibliography file to be used.

\bibliographystyle{ACM-Reference-Format} 
\bibliography{references, zachs_references}

%%%%%%%%%%%%%%%%%%%%%%%%%%%%%%%%%%%%%%%%%%%%%%%%%%%%%%%%%%%%%%%%%%%%%%%%

\end{document}